\documentclass[lettersize,journal]{IEEEtran}
\IEEEoverridecommandlockouts
\usepackage{amsmath,amsfonts}
\usepackage{algorithmic}
\usepackage{algorithm}
\usepackage{array}
\usepackage[caption=false,font=normalsize,labelfont=sf,textfont=sf]{subfig}
\usepackage{textcomp}
\usepackage{stfloats}
\usepackage{url}
\usepackage{verbatim}
\usepackage{graphicx}
\usepackage{cite}
\usepackage{mhchem}
\usepackage{multirow}
\usepackage[dvipsnames]{xcolor}
\usepackage{lipsum} 
\usepackage{hyperref}
\usepackage{xspace}

\makeatletter
\DeclareRobustCommand\onedot{\futurelet\@let@token\@onedot}
\def\@onedot{\ifx\@let@token.\else.\null\fi\xspace}

\makeatother

\begin{document}

\title{Structured Light with Redundancy Codes}

\author{Zhanghao Sun, Yu Zhang, Yicheng Wu, Dong Huo, Yiming Qian, and Jian Wang%
\IEEEcompsocitemizethanks{\IEEEcompsocthanksitem Zhanghao is with the Department
of Electrical Engineering, Stanford University, Stanford, CA 94305 (e-mail: zhsun@stanford.edu)
\IEEEcompsocthanksitem Yu is with the School of Electronic Science and Engineering, Nanjing University, Nanjing 210008, China (email: zhangyu606@gmail.com)
\IEEEcompsocthanksitem Yicheng and Jian are with Snap Research NYC, New York, NY 10036 (email: yicheng.wu@snap.com, jwang4@snap.com)
\IEEEcompsocthanksitem Dong is with Department of Computing Science, University of Alberta, Edmonton, AB T6G 2R3, Canada (e-mail: dhuo@ualberta.ca)
\IEEEcompsocthanksitem Yiming is with the Department of Computer Science, University of
Manitoba, Winnipeg, MB R3T 2N2, Canada (email: yiming.qian@umanitoba.ca)
\IEEEcompsocthanksitem Part of the work was done while Zhanghao was an intern in Snap Research. Jian is the corresponding author 
}% <-this % stops an unwanted space
}

\markboth{Journal of \LaTeX\ Class Files,~Vol.~14, No.~8, August~2021}%
{Shell \MakeLowercase{\textit{et al.}}: A Sample Article Using IEEEtran.cls for IEEE Journals}

% \IEEEpubid{0000--0000/00\$00.00~\copyright~2022 IEEE}

\maketitle

% \IEEEtitleabstractindextext{%
\begin{abstract}
Structured light (SL) systems acquire high-fidelity 3D geometry with active illumination projection. Conventional systems exhibit challenges when working in environments with strong ambient illumination, global illumination and cross-device interference. This paper proposes a
general-purposed technique to improve the robustness of SL by projecting redundant optical signals in addition to the native SL patterns. In this way, projected signals become more distinguishable from errors. Thus the geometry information can be more easily recovered using simple signal processing and the ``coding gain" in performance is obtained. We propose three applications using our redundancy codes: 
(1) Self error-correction for SL imaging under strong ambient light, (2)
Error detection for adaptive reconstruction under global illumination, and (3) Interference filtering with device-specific projection sequence
encoding, especially for event camera-based SL and light curtain devices. We systematically analyze the design rules and signal processing algorithms in these applications. Corresponding hardware prototypes are built for evaluations on real-world complex scenes. Experimental results on the synthetic and real data demonstrate the significant performance improvements in SL systems with our redundancy codes.
\end{abstract}


\section{Introduction}\label{sec:introduction}

% Jian's version, don't delete

\IEEEPARstart{S}tructured light (SL) is a popular activate 3D imaging technique~\cite{gupta2011structured,gupta2013structured}. Similar to a passive stereo system, it recovers 3D geometry via triangulation. Instead of relying on scene appearance for feature matching, an SL system employs a projector to illuminate the scene with encoded patterns and greatly improves effectiveness and efficiency. Typical pattern encoding approaches include time multiplexing, spatial multiplexing, and direct coding.
They have been widely used in various applications, including augmented reality, industrial inspection, and remote sensing.

%,morano1998structured,zhang2006novel,chen2008vision,salvi2010state,feng2019fringe,saragadam2019micro,zhang2008three,

% Current SL techniques can be divided into single-frame and multiple-frame approaches based on the number of projected patterns to acquire a depth map. Single-frame methods continuously project a static pattern to encode the scenarios, which is especially suitable for real-time scanning of dynamic objects, such as the Microsoft Kinect V1 and Intel RealSense~\cite{boyer1987color,freedman2012depth,khoshelham2011accuracy,khoshelham2012accuracy,zhang2012review,sarbolandi2015kinect}. 
%,vuylsteke1990range,je2004high,geng2011structured,geng1996rainbow,
% Multiple-frame approaches require a light emitting device to project a number of patterns so that each projector pixel is temporally encoded.~\cite{posdamer1982surface,geng2011structured,jiang2018three,liu2010dual,saragadam2019micro}. One of the mainstream multiple-frame approaches is the binary coding (gray coding) technique where each projector column is encoded into an unique address~\cite{sansoni1999three,zhang20123,zheng2017phase,wu2019high,wu2019highcyclic,wang2020dynamic}.  In multiple-frame SL, the column/row coordinates of each pixel are encoded through modulation~\cite{srinivasan1984automated,huang2006fast,Garcia2012Consistent,An2016Pixel} 
%liu2010dual,geng2011structured,chen2008modulated,wang2011period,zhang2006high,
% and these scanners can reach an accuracy of tens of microns theoretically~\cite{??}. 

Traditional SL suffers from various challenges, including ambient light, global illumination, and interference.
Strong ambient light significantly decreases the signal-noise-ratio (SNR), limiting SL performance in power-limited mobile devices and 
outdoor environments~\cite{gupta2013structured}. Global illumination, including inter-reflections, diffusion, and sub-surface scattering, breaks the ``direct illumination'' assumption and leads to severe distortions in reconstruction results~\cite{gupta2011structured}. Interference between multiple SL systems or with other active illumination systems creates barriers to separating patterns from different sensors, generating artifacts in the images~\cite{maimone2012reducing}.

% While those challenges are original from the physical system in SL, we find that they also exist in communication (COMM). 
Similar challenges exist in most systems that involve a transmitter and a receiver. One example of special interest in this paper is the digital communication system. Since its invention in the middle 20th, a variety of coding techniques have been proposed and commercialized to boost data transmission fidelity against channel noise, multi-path fading, echos, and multi-user interference. The core idea behind these solutions is to introduce redundancies in signals to rule out certain types of errors. Error detection and correction codes add suffixes to the transmitted data and allow either forward error correction (FEC) or automatic repeat request (ARQ) on the receiver side. Code-division-multiplexing (CDM) technique assigns different codes to multiple users sharing the same communication channel~\cite{cdma}.

% [There are existing solutions...]

% Instead of looking at each challenge based on the specific system design, we \yicheng{treat / analogize} SL as a communication (COMM) system, with a sender (projector) and a receiver (camera). Given this new perspective, we find that solutions in COMM can be adopted into our SL design.

\begin{figure*}[!t]
\centering 
\includegraphics[width=0.8\textwidth]{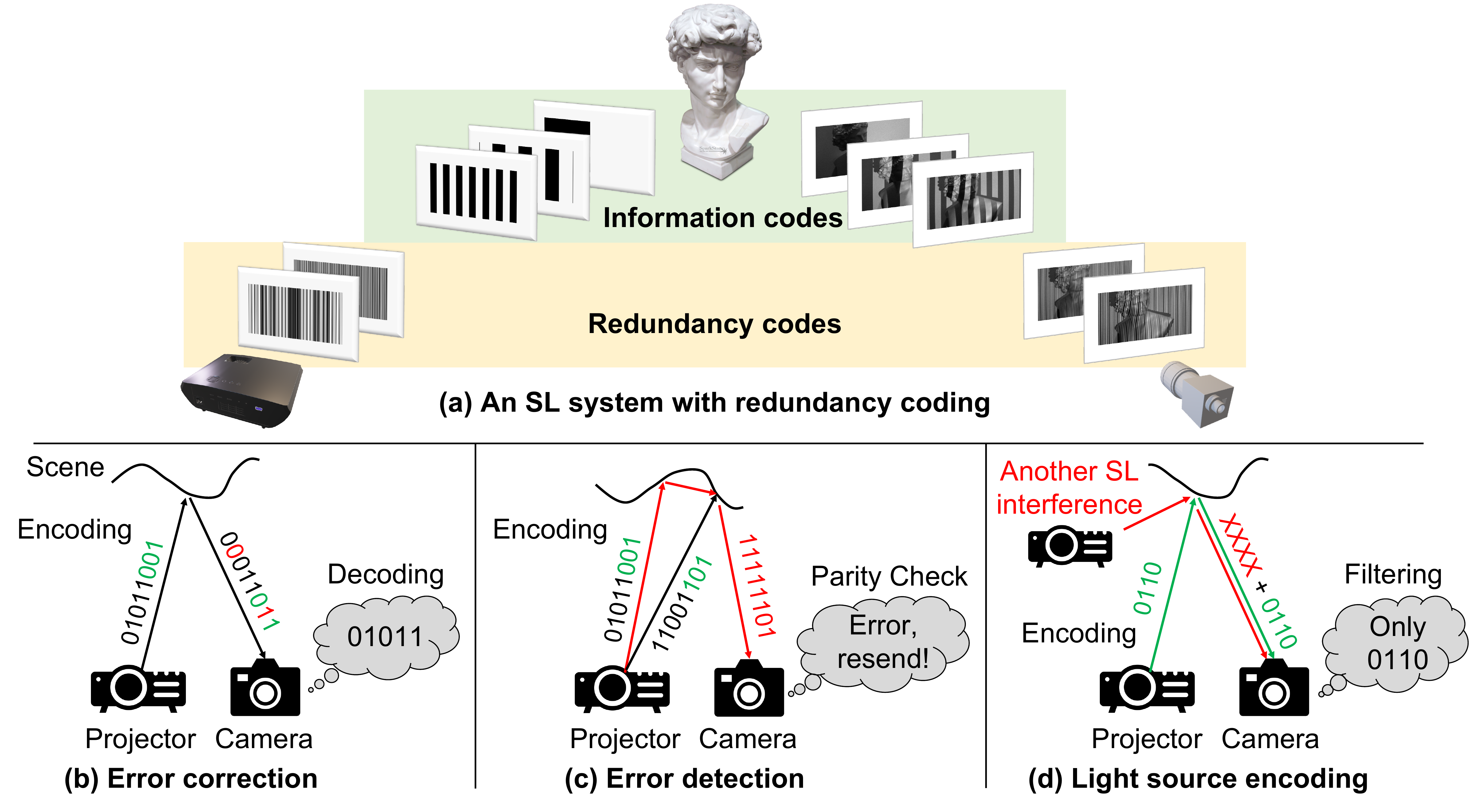}
\caption{(a) A schematic of the proposed SL system with redundancy coding. (b)-(d) Three proposed applications. (b) Error correction. Redundant bits (green) are appended after the data bit string (black) in the encoding process. With this redundancy, a certain amount of errors (red) in the received bit string can be self-corrected in decoding. (b) Error detection. With the redundancies in the transmitted bit string, erroneous data (in this case due to inter-reflection) can be detected through parity check, and another round of (adaptive) measurement is conducted to get more reliable results. (c) Light source encoding. The structured light source only projects a device-specifically encoded temporal sequence to filter out interference from external light sources.}
\end{figure*}

In this paper, we adopt redundancy codes into SL systems to improve their robustness. Firstly, we utilize error detection and correction encoders to improve temporally multiplexed SL. Redundancies are added into the illumination time series at each spatial location to make them more distinguishable from each other. While photon energy in each captured image is reduced due to the redundant frames, a ``\textbf{coding gain}'' is achieved in disparity estimation accuracy because errors can be self-corrected with a maximum likelihood decoder in post-processing. %With the proposed methods, the disparity estimation accuracy is significantly improved under challenging scenarios (e.g., strong ambient light). Another advantage of the redundant coding is that it provides a better measure of the disparity estimation confidence. This confidence can not only serve as starting point in downstream tasks (e.g., sensor fusion~\cite{barron2016fast}, 3D map reconstruction~\cite{izadi2011kinectfusion}), but also help the reconstruction quality.
The effective error correction not only enables accurate geometry reconstruction under challenging scenarios (e.g., strong ambient light) but also provides a reliable measure for the estimation confidence, which typically serves as a useful cue for downstream tasks.

% When the signal degradation goes beyond the error-correction capability, spatial image priors can be applied to borrow information from high confidence estimation and assist the estimation in low confidence regions. 
Secondly, when the degradation level exceeds the error self-correction capability, the system can automatically fall into an error detection mode, where iterative (adaptive) refining measurements are conducted to correct the errors. Lastly, by encoding each structured light ``On'' state into a device-specific code series, the SL system is able to identify its own light from external temporally changing light, either from another SL device or from ambient light. Interference-induced artifacts can be largely removed.

Note that these adoptions are not trivial due to different transmission channel properties and different performance requirements. Essentially, we answer two questions: which hardware system and physical environments allow the most benefits from utilizing redundancy codes? and how the coding technique can be integrated with insights specific to SL systems? Compared with prior methods that focus on special hardware or ad-hoc coding mechanism, our approach is a more general plug-and-play solution. It can serve as a supplement to previously proposed SL techniques that further increases their robustness.

We summarize our contributions in the following.
\begin{itemize}
    \item We adopt redundancy coding techniques to SL systems as a low-cost and general-purposed approach against strong noise and interference. 
    \item We analyze the benefits and limitations of these techniques specific to SL, and further propose corresponding design rules and post-processing algorithms.
    \item We build hardware prototypes to demonstrate the effectiveness of these approaches in different SL systems. Compelling results are achieved in both synthetic and real experiments.
\end{itemize}

\textbf{Limitations}. The major limitation of using redundancy coding in SL is the increase in the number of image frames to be captured. This either requires a higher frame rate image system or a longer data acquisition time. However, in this paper, we focus on applications where frame number increase is small (e.g., less than $2\times$) such that it will not pose severe challenges to hardware capability. Also, the proposed design rules aim at using as few redundancies as possible to achieve the desired performance gain.

% \zh{Leave to Jian} \jian{need to acquire more frames. But here same total acquisition time. Can be deleted} Needs higher frame rate (we provide choices of difference encoding for different frame rate systems \yicheng{will this "different frame rate systems" confict with the point that it does not require special hardware. Need to be careful}\zh{perhaps we just remove this. Since we just have about 2 times more frames, and the this .} ). 

% end of Jian's version

% end of Jian's version

\section{Related work}
\subsection{Structured light encoding methods}
The SL system
can be classified into three categories based on encoding approaches \cite{salvi2004pattern}: (1) time-multiplexing, (2) spatial-multiplexing, and (3) direct-coding.

Time-multiplexing methods project a sequence of patterns in a short period of time and captures a series of images. Typical time-multiplexing methods include binary code \cite{posdamer1982surface}, gray code \cite{inokuchi1984range}, phase shifting \cite{bergmann1995new}, and several hybrid methods \cite{hall2001stripe}. 
Spatial-multiplexing methods exploit spatial neighbourhood to encode the scenes, including M-arrays \cite{morita1988reconstruction}, random dots pattern\cite{sarbolandi2015kinect}, and non-formal codification \cite{maruyama1993range}. Direct-coding methods utilize intensity or colour to encode the pixel directly, such as codification based on grey levels \cite{hung19933d} and codification based on colour \cite{tajima19903}. Time-multiplexing related methods can provide accurate, high spatial resolution reconstruction. 
Efforts have been made to optimize the number of projected patterns~\cite{Mirdehghan_2018_CVPR, gupta2018geometric}. On the other hand, most consumer SL devices project spatially multiplexed random dot patterns with compact and lightweight diffractive optical elements.~\cite{zhang20173d}.

\subsection{SL challenges and existing solutions}
%To anti-sunlight,  and scan, etc.
%See my Survey paper in Chinese.

%To anti-global light, Mohit's paper \cite{gupta2011structured} designs codes for global illumination.

%To anti-device-device interference,

Structured light-based techniques usually face challenges under complex environments, such as ambient light, global illumination, and multi-path interference. Previous methods mostly focus on complex hardware or ad-hoc projection pattern designs, while our redundancy coding solution is more general-purposed and can be easily integrated with the existing techniques for further performance enhancement.

\noindent \textbf{Ambient light:}
Strong ambient light (e.g., Sunlight) severely degrades the reconstruction performance of SL. Researchers have proposed methods employing special hardware such as bandpass filters~\cite{padilla2005advancements}, polarizers~\cite{padilla2005advancements}, and near infrared systems~\cite{wang2016dual}, as well as  optimized scanning mechanisms~\cite{ gupta2013structured,bartels2019agile}.

% Gupta~{\it et al.}~\cite{gupta2013structured} propose a method of light-concentration to overcome strong ambient light, especially in strong sunlight in outdoor scenarios. Mertz~{\it et al.}~\cite{mertz2012low} demonstrate a solution to utilize the projector’s high-speed MEMS mirror motion and laser light-sources for suppressing ambient light in order to realize low-cost and low-power reconstruction of outdoor scenes in sunlight. \jian{not good. Mertz is not a good paper. Refer to my Chinese survey}

\noindent \textbf{Global illumination:}
In complex scenes~\cite{GuptaA}, direct and indirect/global illumination are usually mixed together. In order to separate the direct light path from the indirect illumination/global illumination, Nayar {\it et al.}~\cite{nayar2006fast} proposed a fast and effective method by projecting a high frequency pattern and its complement. Follow-up techniques including adaptive correspondence~\cite{xu2007robust}, multiplexed illumination~\cite{gu2011multiplexed}, modulated phase-shifting~\cite{chen2008modulated}, modified gray code~\cite{gupta2011structured}, micro phase-shifting~\cite{gupta2012micro}, and
embedded phase-shifting~\cite{moreno2015embedded} are also proposed.

% Gupta~{\it et al.}~\cite{gupta2011structured} analyze the errors caused by global illumination in gray codes SL and propose novel gray code that is resilient to global illumination. 
% O’Toole~{\it et al.}~\cite{o2014temporal} modify the optical system to separate the direct and indirect light paths by utilizing the epipolar geometry constraint to block the global illumination during data acquisition.

\noindent \textbf{Multi-device interference:}
When multiple SL devices are present in the same environment, the projected pattern from one device will interfere with the other ones, leading to strong artifacts in geometry reconstruction. To solve this problem, methods are developed using motion~\cite{maimone2012reducing}, multiple illumination wavelengths (colors)~\cite{cronie2019coordination}, and defocus blur~\cite{wu2020freecam3d}.
Recently, multi-path interference caused by a camera pixel collecting more than one light path from either multiple devices~\cite{maimone2012reducing}, or multiple pixels in a single device~\cite{zhang2019causes, zhang2021sparse}, is receiving extensive attention from both academia and industry.

% To address the multi-path interference caused by a single structured light setup, Zhang~{\it et al.}~\cite{zhang2019causes} model the issue of camera pixel receives light from exactly two positions with a bimodal multi-path model and provide an effective approach to separate the two light paths with a two-step phase reconstruction method. 
% Zhang~{\it et al.}~\cite{zhang2021sparse} examine the sparse or N-modal multi-path phenomenon and present a solution by treating each camera pixel as an underdetermined linear system of equations and finding the sparsest solution via an application-specific Bayesian
% learning approach.
% To reduce interference between two structured light sensors, Maimone~{\it et al.}~\cite{maimone2012reducing} apply a small amount of
% motion to one sensor so that each sensor can observe a clear and sharp pattern from itself, while a blurry pattern from the others.
% Combine SL with defocus cue~\cite{wu2020freecam3d}

\subsection{Redundancy codes in Communication}
Both communication and SL systems can be abstracted into a transmitter/receiver architecture. In digital communication systems, channel noise, multipath propagation, and interference~\cite{comm_textbook} impairs reliable data transmission. Redundancy codes are proposed as an effective solution. Error correction and detection codes used in communication systems include the conventional linear block codes~\cite{bch_code1, golay_code} and more advanced codes such as turbo code~\cite{turbo} and LDPC code~\cite{ldpc}. Redundancy codes are also utilized to achieve user channel multiplexing. One of the most successful examples is the code-division-multiplexing (CDM) technique for 3G network~\cite{cdma}. Similar techniques are also employed in data storage, barcodes, and Radar/LiDAR sensing~\cite{comm_textbook, lidar_cdma}.

\subsection{Redundancy codes in SL}
The idea of utilizing redundancies in SL has been explored in a few prior arts.
% Gupta~{\it et al.}~\cite{gupta2011structured} use four groups of codes to handle a different range of inter-reflections \textcolor{red}{[zh: the major focus of gupta's paper is not redundancy, but ad hoc patterns?]}
% In this paper, we just append more bits to each projector pixel to involve redundancy and facilitate adaptive scanning procedure.
Sagawa~{\it et al.}~\cite{sagawa2017illuminant} propose to use direct sequence spread spectrum (DSSS) under strong ambient light. The encoding process requires $ 15 \sim 200 \times$ more frames, which can only be applied in a high-speed SL system (with $>10$kFPS).
%is a more general approach and can be used in a common time-multiplexing SL system without any hardware modification.
Li~{\it et al.}~\cite{li2021error} proposes an error self-correction method in phase-shifting SL by involving redundant measurement to address the phase jumps at the edges of the fringes. Rosario~{\it et al.}\cite{porras2017error} propose the use of N-ary gray level Reed-Solomon codes to improve the reliability against noise. However, both phase-shifting and N-ary gray level SL fail under strong ambient light. A detailed discussion and simulation are provided in supplementary material. In this paper, we focus on using redundancy coding techniques to improve SL robustness in very challenging environments.

\section{SL with error correction codes}
% Jian's version, don't delete
\label{sec:ecc}
\subsection{Principle of error correction codes (ECC)}
\label{sec:ecc_principle}
\begin{figure}[!t]
\centering 
\includegraphics[width=1.0\columnwidth]{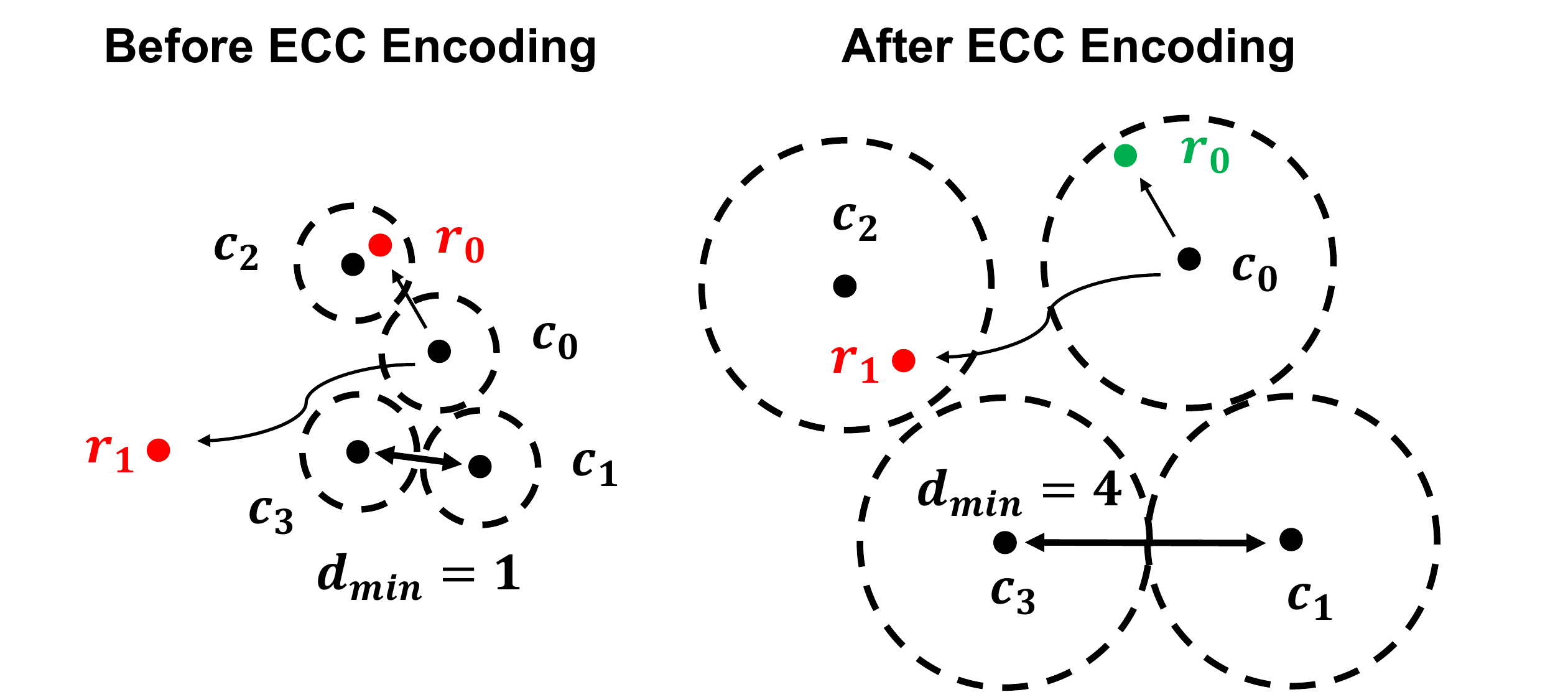}
\caption{Error correction encoding and decoding processes with the codebook $c_0 \sim c_3$. Before ECC Encoding, $d_{min} = 1$, and none of $r_0$, $r_1$ can be decoded correctly. After ECC encoding, $d_{min}$ increases to $4$ and $r_0$ is decoded correctly.}
\label{fig:dmin}
\end{figure}

% The distance disk
We first give a brief summary of the basic ECC principle. For comprehensive reviews on this topic, please refer to ~\cite{ecc_textbook, comm_textbook}. Suppose the transmitter sends out a set of data bit-strings $C = \{\mathbf{c}\}$, noise and interference in the data transmission process lead to errors on the receiver side. If the noise is relatively small, such that the received bit-string $\mathbf{r}$ only slightly deviates from the original bit-string $\mathbf{c_0}$, the receiver is able to retrieve $\mathbf{c_0}$ by searching for the data string that is most similar to $\mathbf{r}$ in the codebook $C$, as shown in the green scenario in the right side of Fig.~\ref{fig:dmin}. This ``decoding'' (error self-correction) process is only valid when the original string $\mathbf{c_0}$ is closer to $\mathbf{r}$ than any other strings in the codebook. Otherwise, as shown in the red scenario on the rights side of Fig.~\ref{fig:dmin}, the decoding leads to an incorrect estimation $\mathbf{c_2}$.

Therefore, if the codebook has bigger discrepancies between any two codes in it, the decoding is more robust. This is denoted as \textbf{coding gain} in communication systems. The discrepancy between two bit-strings is usually defined as the Hamming distance (number of different symbols). 
% This is similar to the $L_0$ distance in computer vision techniques. Note that in the context of structured light systems, using other distance metrics (e.g., $L_1$, $L_2$) does not make a difference.\jian{?}\zh{need to check}
$d_{min}$ is defined as the minimum distance between any two codes in the codebook, as shown in Fig.~\ref{fig:dmin}. The decoding process is guaranteed to succeed if the number of erroneous bits is less than $d_{min}/2$, and a codebook is usually represented by the tuple ($n$, $k$, $d_{min}$). 

For a codebook of $k$ bit-strings, there can be mostly $2^k$ different data strings. $d_{min}$ of this non-redundant codebook is only $1$ (left part of Fig.~\ref{fig:dmin}). To increase $d_{min}$, ECC encodes each $k$ bit-string into an $n$ bit-string ($n > k$), with $n-k$ redundant bits. Generally, with larger $n$, it is easier to get larger $d_{min}$ and higher error correction capability. With $n$, $k$ fixed, $d_{min}$ is upper bounded by $n-k+1$~\cite{ecc_textbook}. Theoretically, such codebook can be generated with a brute force search in $n$-dimensional space. However, the search becomes prohibitively expensive even with relatively small $n$ and $k$ (e.g., $\sim 10$) (we include one such example in the supplementary material). Fortunately, several families of high performance error correction codes (with large $d_{min}$) have been designed~\cite{golay_code, bch_code1, rm_code}. In the following section, we elaborate on adopting these codes to SL systems.
% In this paper, we aim at adapting these codes and corresponding encode/ decoders to SL systems.

\subsection{Error correction code in SL}
% Here we and elaborate on the key differences between SL and COMM systems
\subsubsection{Differences between SL and communication}
\label{sec:difference}
Although both SL and communication systems can be abstracted into the transmitter-receiver architecture, two key differences exist. First, most communication systems are specifically designed for large data blocks transmission, with $k \sim 1000$ or more, while in SL systems, each bit-string is usually shorter than $10$bits. This eliminates the concern of computational cost and makes brute force encoding/decoding processes possible. Second, noise in most communication channels is signal-independent additive white Gaussian noise (AWGN). However, the dominant source of noise in SL is signal-dependent photon shot noise from the projector light and ambient light. Therefore, the effective regime is different: in communication systems, ECC is usually effective with error rates less than $\sim 1\%$ \footnote{While this is a small amount of error for most computer vision tasks, this regime is critical for communication systems}. On the other hand, we demonstrate the effectiveness of ECC with error rates $\sim 10\%$ to $70\%$ in the SL system.

\subsubsection{SL image formation model}
\begin{figure*}[!t]
\centering 
\includegraphics[width=0.9\textwidth]{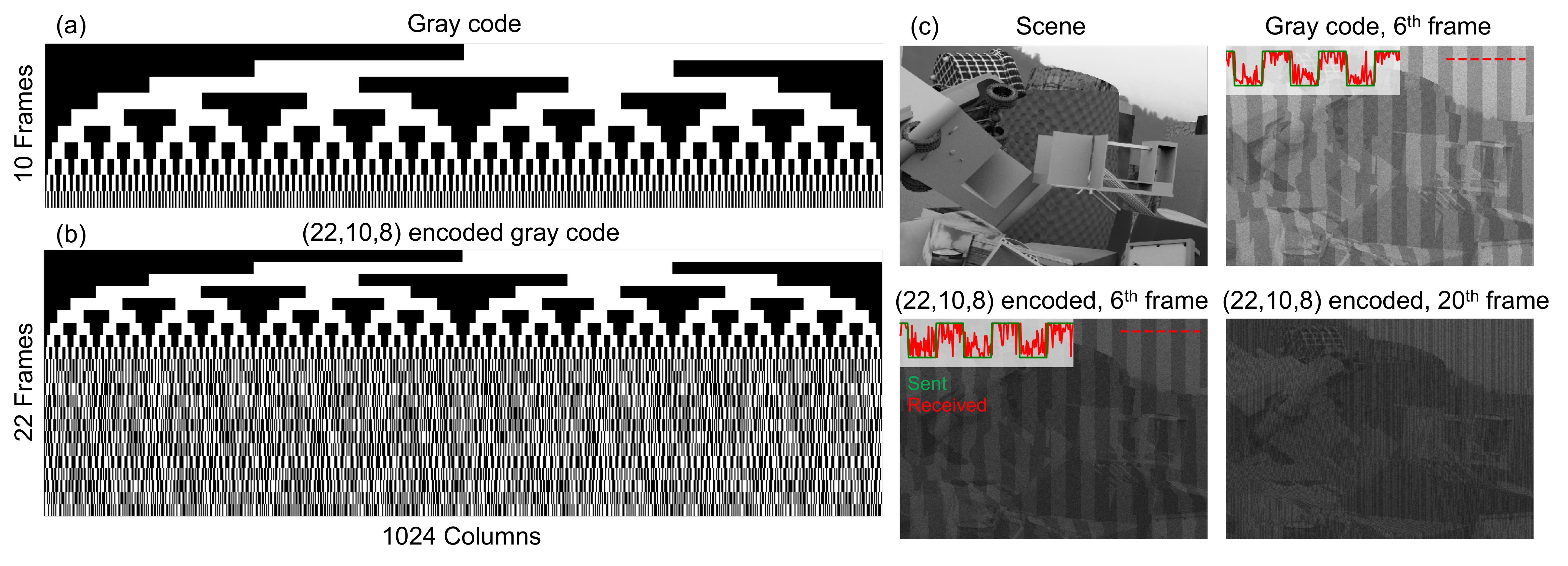}
\caption{Synthetic data for method evaluation. (a) Original 10bit gray code (b) ($n = 22, k = 10, d_{min} = 8$) encoded 10bit gray code, 12 ECC bits are appended to each 10bit string. (c) Scene and captured images. Compared to conventional SL, ECC SL projects more frames within a fixed total exposure time, and the per-frame SNR is lower. We also show normalized cross-sections of the sent (green) and received (red) signals along the dashed line. All frames are brightened by $3\times$ for visualization.} 
\label{fig:simu_data}
\end{figure*}

In a conventional binary or gray code (GC) SL, to acquire a depth map with $M$ columns, $k = \textrm{log}_2 M$ pattern frames $\{\mathbf{P}_i\}$ need to be projected. Each $\mathbf{P}_i$ is a binary map with $0$s and $1$s. $M$ is determined by the projector's resolution, baseline between projector and camera, and the scene dimensions of interest. Typically, $8 \sim 10$ frames are used to get $256 \sim 1024$ depth map resolution. In the hybrid phase-shifting/binary pattern SL, a smaller amount of column numbers are projected (e.g., $3 \sim 5$) to alleviate the phase ambiguity. As shown in Fig.~\ref{fig:simu_data} (a), each column in the projected pattern sequence is an unique $k$bit string (in temporal dimension). These pattern frames form a $k\times N \times M$ data cube, where $N$, $M$ are the row and column resolutions. Projecting these pattern frames is equivalent to sending $N\times M$ $k$-bit data strings in a communication system, where scene geometry information is encoded. 
On the camera (receiver) side, the image formation model can be expressed as
\begin{eqnarray}
\label{eqn:noise_model}
    \mathbf{I}_i = \mathcal{W}[\mathbf{P}_i] \odot \mathbf{A} + \mathbf{I}_a + \mathcal{N}
\end{eqnarray}
Where the captured image frames $\{\mathbf{I}_i\}$ form another $k\times N \times M$ data cube. $\mathcal{W}$ is the depth-dependent warping process. $\odot$ is the pixel-wise multiplication operation. $\mathbf{A}$ is the scene texture, determined by material properties and surface normal. $\mathbf{I}_a$ is the ambient light signal. $\mathcal{N}$ includes photon shot noise, sensor readout noise, and quantization error. We use the heteroscedastic Gaussian noise model to account for both readout noise and shot noise. For image signal $\mathbf{I}$ normalized to $\in [0,1]$, the variance of $\mathcal{N}$ can be expressed as $\sigma_r^2 + \sigma_s^2\mathbf{I}$. Under normal lighting conditions and scene albedo settings, photon shot noise dominates. However, when the overall illumination level is low, or when the scene albedo is low, sensor readout noise and quantization error dominate. 

Disparity estimation process is to find the best match between each bit-strings in the projector pattern and captured image data cubes in the spatial domain. In a rectified SL system, this search can be conducted on each epipolar line. Once the correspondence is established $\mathbf{I}[n,m] \leftrightarrow \mathbf{P}[n, \widehat{\mathbf{corr}}[n,m]]$ (where $\widehat{\mathbf{corr}}$ is the estimated correspondence map), disparity is obtained as $\widehat{\mathbf{disp}}[n,m] = m - \widehat{\mathbf{corr}}[n,m]$. From the disparity, the 3D geometry of the scene is reconstructed. 

\subsubsection{Encoder design} 
\label{sec:encoder_design}
In ECC SL systems, the original $k$ data frames are encoded into $n$ frames ($n > k$). With a fixed total exposure time, per-frame exposure time is $k/n$ times shorter, and $n$ is upper bounded by the projector (usually $30$/$60$ FPS) and the camera frame rates ($\sim 100$ FPS). Also, with a shorter exposure time, the signal-to-noise-ratio (SNR) is lower for each frame. Therefore, although bigger $n$ usually provides a larger $d_{min}$, it is not always preferred. We show one such example in Sec.~\ref{sec:simu_results}.

We examine several high performance code families, including (extended) binary Golay code~\cite{golay_code}, Reed-Muller code family~\cite{rm_code}, and BCH code family (Goppa code family)~\cite{bch_code1}, and summarize their coded bit-string length $n$, data bit-string length $k$, and minimum Hamming distance $d_{min}$ in a lookup table in the supplementary material. One example is shown in Fig.~\ref{fig:simu_data} (b). The $(22,10,8)$ code appends $12$bits after the $10$bit uncoded data string (Fig.~\ref{fig:simu_data} (a)), and the codebook has minimum distance $d_{min} = 8$.

Similar to the conventional SL patterns, the arrangement of columns in ECC SL patterns also matters. Less `0-1' jumps in the pattern reduce artifacts in real-world experiments~\cite{geng2011structured}. For a detailed discussion, please refer to the supplementary material.

% In ours, ($n = k+r$, $k$, $d$) ($r>0$). Usually, an SL system has an upper bound on the frame rate, due to projector (designed to be 30fps or 60fps) and camera (100fps for a machine vision camera, 30fps or 60fps for a web camera), then use the upper bound frame rate. In practice, $k$ is decided by: the number of columns of the projector, the size of the target, the maximum disparity of the SL system, the baseline between the camera and projector, the depth range of interest. 

% Now the task is given $n$ which corresponds to the upper bound of the frame rate, find $2^k$ binary vectors, such that the minimum Hamming distance is maximized.

% Method 1: random search. 

% Method 2: based on existing codes: Hamming code, BCH code, truncated BCH code, Reed–Muller code, Golay codes, convolutional codes, punctured convolutional codes, Goppa code. The Maximum Distance Separable (MDS). The Singleton Bound. But binary code which meets the Singleton Bound doesn't exist.

\subsubsection{Decoder design} 
As discussed in Sec.~\ref{sec:ecc_principle}, the decoding process can be formulated as
\begin{eqnarray}
\label{eqn:decode}
    \widehat{\mathbf{c}} = \arg\min_{\mathbf{c} \in C} d(\mathbf{c}, \mathbf{r})
\end{eqnarray}
Where $\mathbf{r}$ is the received signal string, $\widehat{\mathbf{c}}$ is the estimated (decoded) bit-string, $C$ is the codebook, and $d(,)$ is a distance metric. Since the size of the codebook $C = 2^k$, in communication systems with $k \sim 100-1000$, it is impossible to traverse the codebook $C$ and find the best match. $\mathbf{r}$ is usually first binarized and fed into an ad hoc decoder circuit based on the specific ECC algebraic properties. This process can be sub-optimal and throws away information in the quantization pre-processing. In SL systems, brute force correspondence search can be conducted in the floating-point number domain without any loss of information. This process is sometimes referred to as ``soft-decision-decoding'' in communication systems, in contrast to the sub-optimal, binarized ``hard-decision-decoding'' algorithm.
% With the noise model defined in Eq.~\ref{eqn:noise_model}, the decoding problem in Eq.~\ref{eqn:decode} is expressed as an maximum-likelihood-estimation (MLE) problem.

With the self-correction capability in ECC SL, large portions of errors are avoided in the decoding process. However, in low SNR situations, errors in the estimated disparity map still exist. A more advanced technique named ``list decoding'', is applied in modern communication systems~\cite{list_decode1}. Its core idea is to keep several candidates in the decoding process and make a decision later based on certain priors. This is similar to the image restoration algorithms in computer vision. Information beyond a single pixel is utilized to assist the processing. Different from most image denoising tasks, in an SL system, errors are dominated by impulse noise. As an example, a one-bit error in the decoding process can lead to disparity error $ = 1$ or disparity error $=1000$ with the same probability. The median filter is commonly utilized due to its robustness to outliers~\cite{med_filter}. However, with the same weight assigned to each pixel in the median calculation, the median filter generates clustered artifacts under low SNR conditions. Therefore, it is important to identify which estimation is reliable before employing external information.

Thanks to the more distinguishable codes, ECC SL can provide more reliable confidence estimations. We define the confidence as $(d_2 - d_1)/d_2 \in [0,1]$, where $d_1$, $d_2$ are the smallest and second smallest distances between codes in the codebook and the received signal $\mathbf{r}$. Based on the confidence map, we developed two simple algorithms to further reduce disparity estimation errors. First, a confidence-based median filter that only includes high confidence pixels in the median calculation is implemented. It utilizes external information within a local patch around each low confidence pixel. Two thresholds $t_{low}$ and $t_{high}$ are used to classify one pixel to be low confidence or high confidence. Second, an ``order prior'' is introduced to utilize more global external information. It is based on the observation that the order of columns in projected and warped patterns does not change for continuous surfaces. In the decoding, we select high confidence ``anchor points'' along each epipolar line and check the order relationship between these anchor points and the low confidence points. If the order relationship does not hold, a correspondence candidate $\mathbf{c}$ with larger (sub-optimal) distance $d(\mathbf{c}, \mathbf{r})$, while satisfying the order relationship, is selected. Note that there are edge cases for both confidence-based median filter and the order prior, especially in complicated geometry (e.g., a fine grid). Also, if the confidence estimation is incorrect, erroneous pixels might be selected as ``anchor points'' and worsen the reconstruction result. This situation is much less likely in the ECC SL compared to conventional SL.

\subsubsection{Simulation results}
\label{sec:simu_results}
We evaluate the encoding and decoding processes on synthetic FlyingThings3D dataset~\cite{MIFDB16}. We synthesize noisy captured images based on the image formation model Eq.~\ref{eqn:noise_model}. Two calibration images (one with projector light all on and one with projector light all off) are additionally captured. Total exposure time is fixed for all methods. Fig.~\ref{fig:simu_data} (c) shows one data frame from original gray code SL, one data frame from a $(22, 10, 8)$ ECC SL, and an error correction frame. It can be seen that the ECC SL frames are dimmer and noisier due to the shorter per-frame exposure time.

\begin{figure}[!t]
\centering 
\includegraphics[width=0.7\columnwidth]{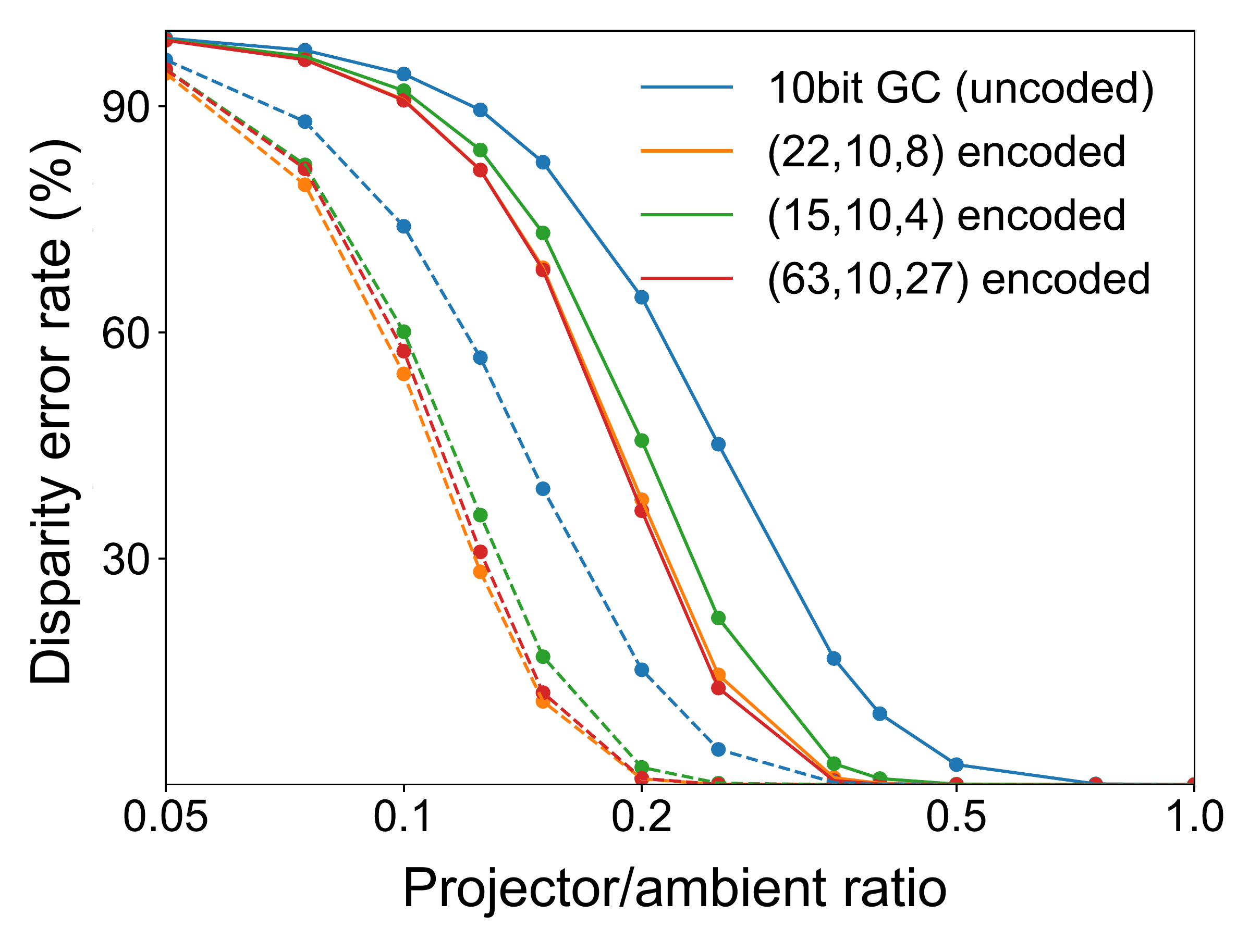}
\caption{Quantitative comparisons between the original GC and ECC SL, under strong ambient light and shot noise dominant situation. Solid lines and dashed lines are with different photon shot noise parameter settings. The trends are the same for the two settings.}
\label{fig:soft_quant}
\end{figure}

\begin{figure}[!t]
\centering 
\includegraphics[width=1.0\columnwidth]{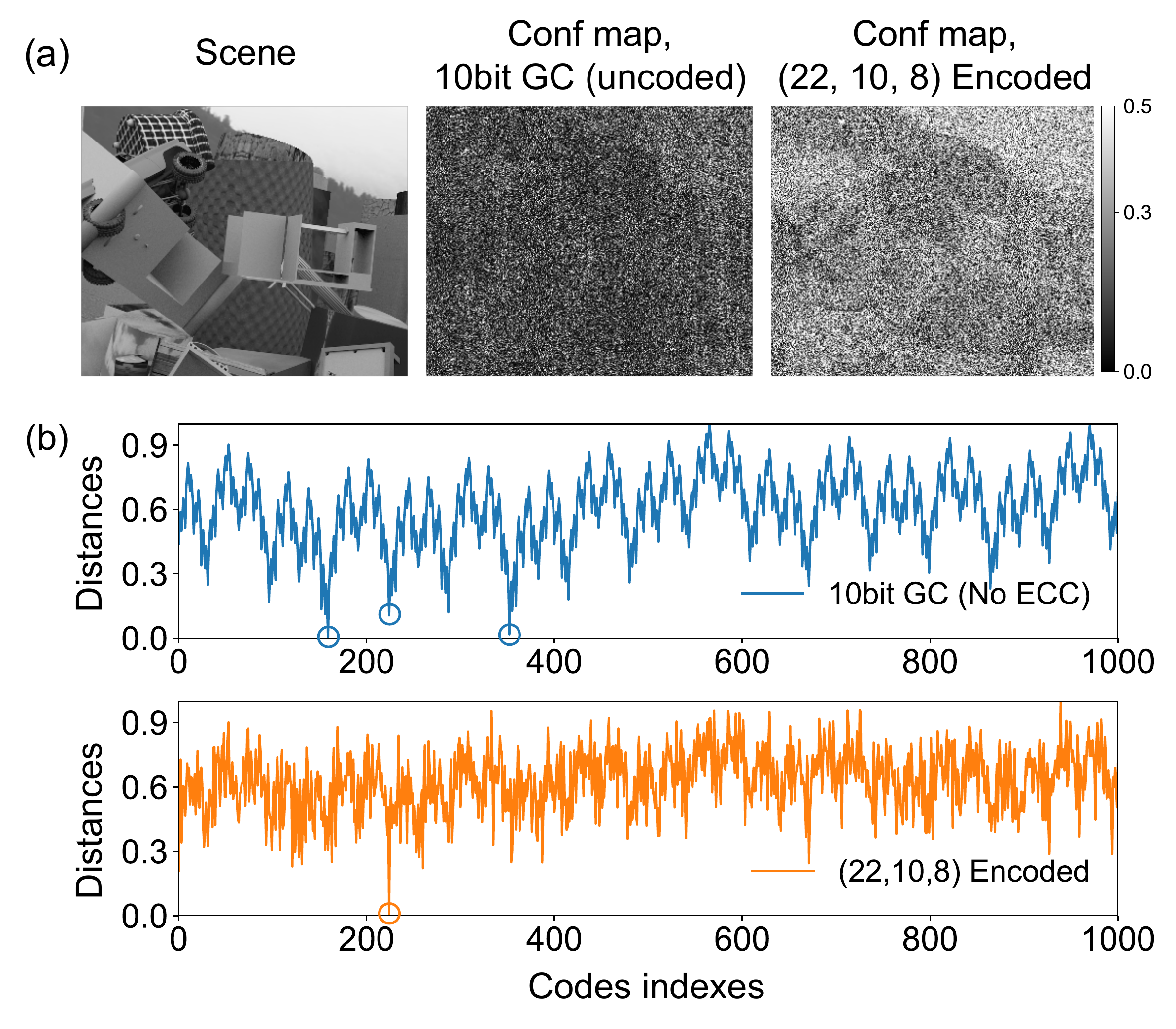}
\caption{Confidence map from original GC SL and ECC SL. (a) Qualitative comparison between the two confidence maps. (b) Correspondence search process at one pixel with and without ECC. Empty circles indicate possible correspondences $\mathbf{c}$ with small distances $d(\mathbf{c}, \mathbf{r})$. Both distances are normalized to $[0,1]$}
\label{fig:conf_map}
\end{figure}

\begin{figure}[!t]
\centering 
\includegraphics[width=0.7\columnwidth]{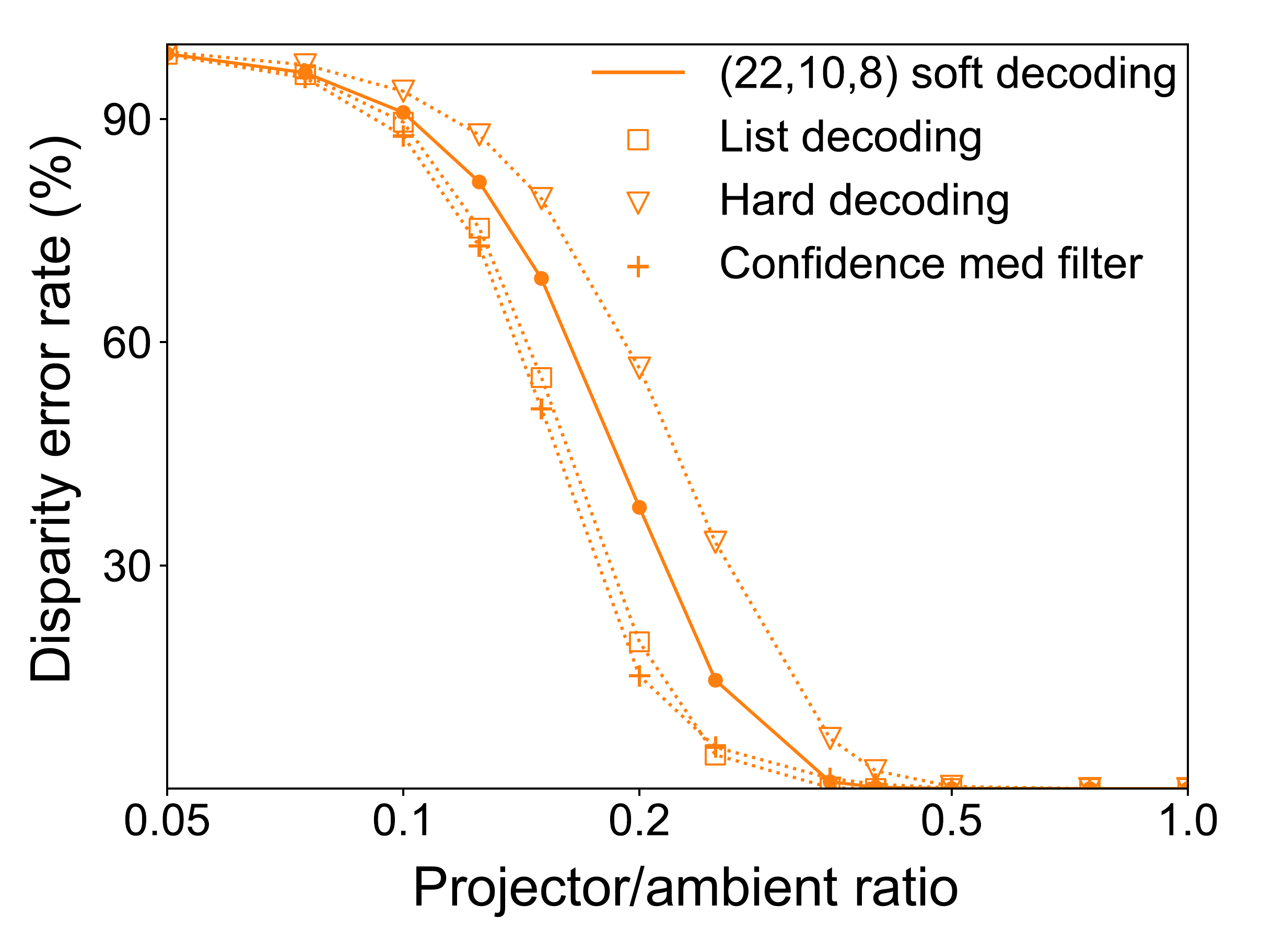}
\caption{Quantitative comparisons between different decoding methods.}
\label{fig:decode_quant}
\end{figure}

\begin{figure*}[!t]
\centering 
\includegraphics[width=0.9\textwidth]{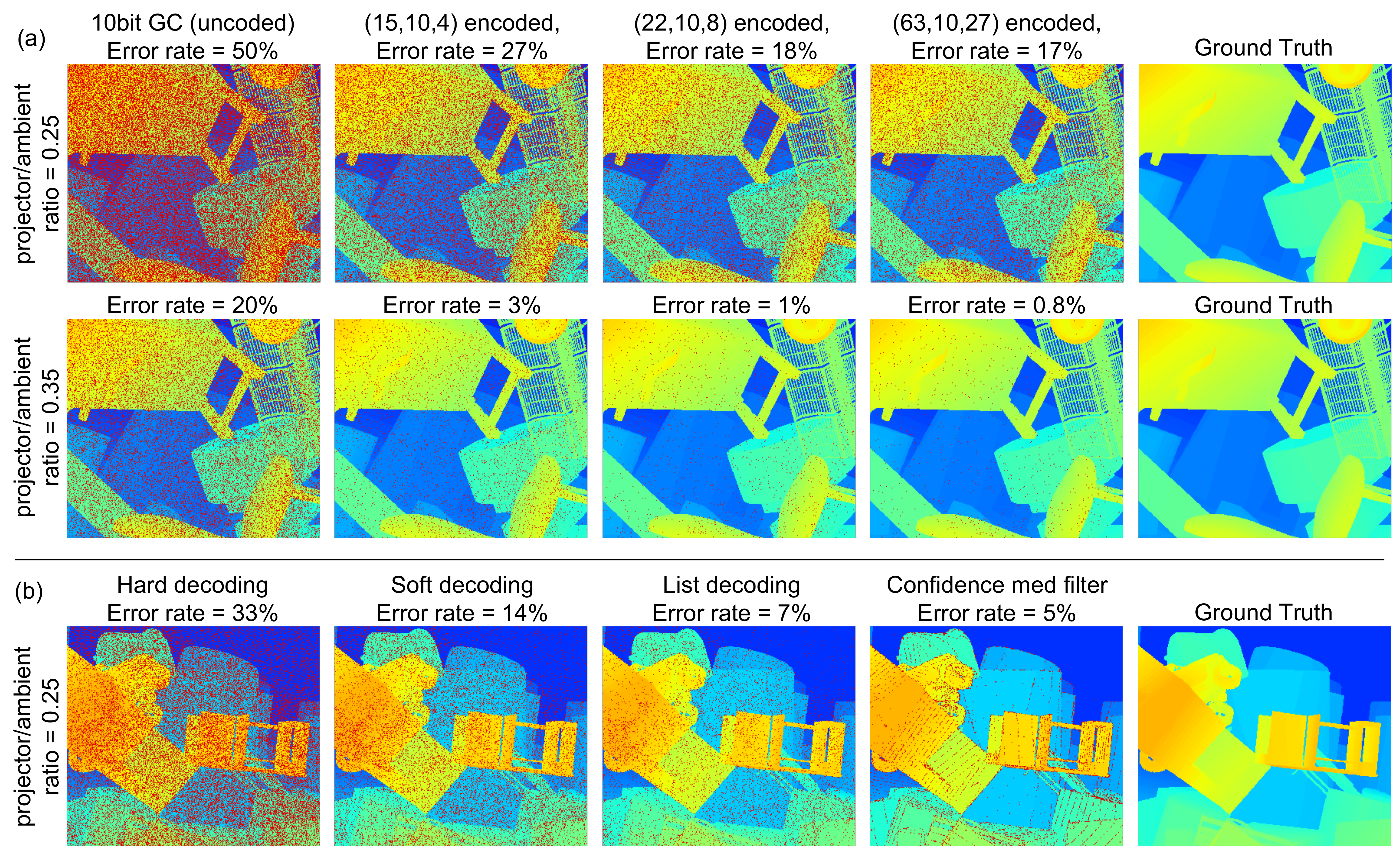}
\caption{Qualitative comparisons on SL performance (synthetic data). All error pixels are marked out by red color. (a) Comparisons between original GC code and different ECC, under different ambient light intensity settings. ECC SL consistently achieves higher quality reconstruction compared to conventional SL. (b) Qualitative comparisons between different decoding methods.}
\label{fig:soft_qual}
\end{figure*}

The quantitative comparisons on soft-decision decoding results are shown in Fig.~\ref{fig:soft_quant}. Here we compare four SL systems: the original $10$bit gray code (GC) SL, and three ECC SL with $(15,10,4)$, $(22,10,8)$, $(63,10,27)$ codebooks. We define error rate as the percentage of pixels with incorrect disparity estimations. We fix the camera parameter settings, and tune the ratio between projector illumination intensity and ambient light intensity. It can be seen that when SNR is extremely low (projector light $\ll$ ambient light, error rate $\sim 100\%$) or extremely high (projector light $\gg$ ambient light, error rate $\sim 0\%$) , coding does not give much benefits, which is within expectation. However, within the broad range between these two extremes, coding reduces the error rate by $3\times$ or more. We show the error rate curves with two camera shot noise parameter $\sigma_s$ settings: $\sigma_s = 0.015$ (dashed) and $\sigma_s = 0.04$ (solid). It can be seen that both curves show similar trends, except a horizontal shift. Note that the long $(63,10,27)$ code only provides marginal performance increase compared to the $(22,10,8)$ code. Therefore, it should not be used in real-world ECC SL even if it has very large $d_{min}$.

There is a special scenario where redundant coding is not beneficial, or even impairs accuracy. That is, when the received projector light is extremely weak, due to low albedo in the scene, weak illumination power, or long imaging distance (e.g., SL in endoscopy system~\cite{sl_encodscopy}). In this regime, sensor readout noise and quantization error dominates. Please refer to supplementary material for details.

Confidence maps from conventional SL and ECC SL are shown in Fig.~\ref{fig:conf_map} (a). It can be seen that in the original 10bit GC SL, most pixels have low confidence, while in the ECC SL, high confidence pixels are much more. A comparison between the correspondence search processes with and without ECC at one pixel is shown in Fig.~\ref{fig:conf_map} (b). In the conventional SL system, correspondence search achieves small distances for multiple candidates. Thus, the confidence is low. However, in ECC SL, only one optimal matching result is achieved, with distance $d(\mathbf{c}, \mathbf{r})$ much smaller than any other codes in the codebook. By utilizing the reliable confidence map and local/global priors in the SL system, list decoding and confidence-based median filtering are able to further reduce the error rate in ECC SL from soft-decision decoding. 

We then compare different decoding algorithms in ECC SL quantitatively: soft-decision decoding, hard-decision decoding, list decoding with order prior, and confidence-based median filtering. We use the $(22,10,8)$ code with soft-decision decoding as the baseline. As shown in Fig.~\ref{fig:decode_quant}, hard-decision decoding utilizes computationally efficient generic ECC decoders. However, it leads to a higher error rate due to the loss of information in the binarization. Note that since most computations in the decoding process occur in correspondence search between received signals and codes in the codebook, an efficient GPU matrix multiplication implementation can achieve $>200$fps for $600\times700$ spatial resolution.
% an approximate nearest neighbor (ANN) search algorithm can be utilized to increase processing speed by $\sim 8\times$. It also supports outputting multiple candidates in the search process, which is useful in list decoding. We show detailed performance analysis in the supplementary material. 
In the list decoding, we keep $3$ peaks from the first-stage correspondence search. In the confidence-based median filtering, we use a $5\times 5$ window size. Both methods further increase the decoding accuracy significantly.

\begin{figure*}[t]
\centering 
\includegraphics[width=1.0\textwidth]{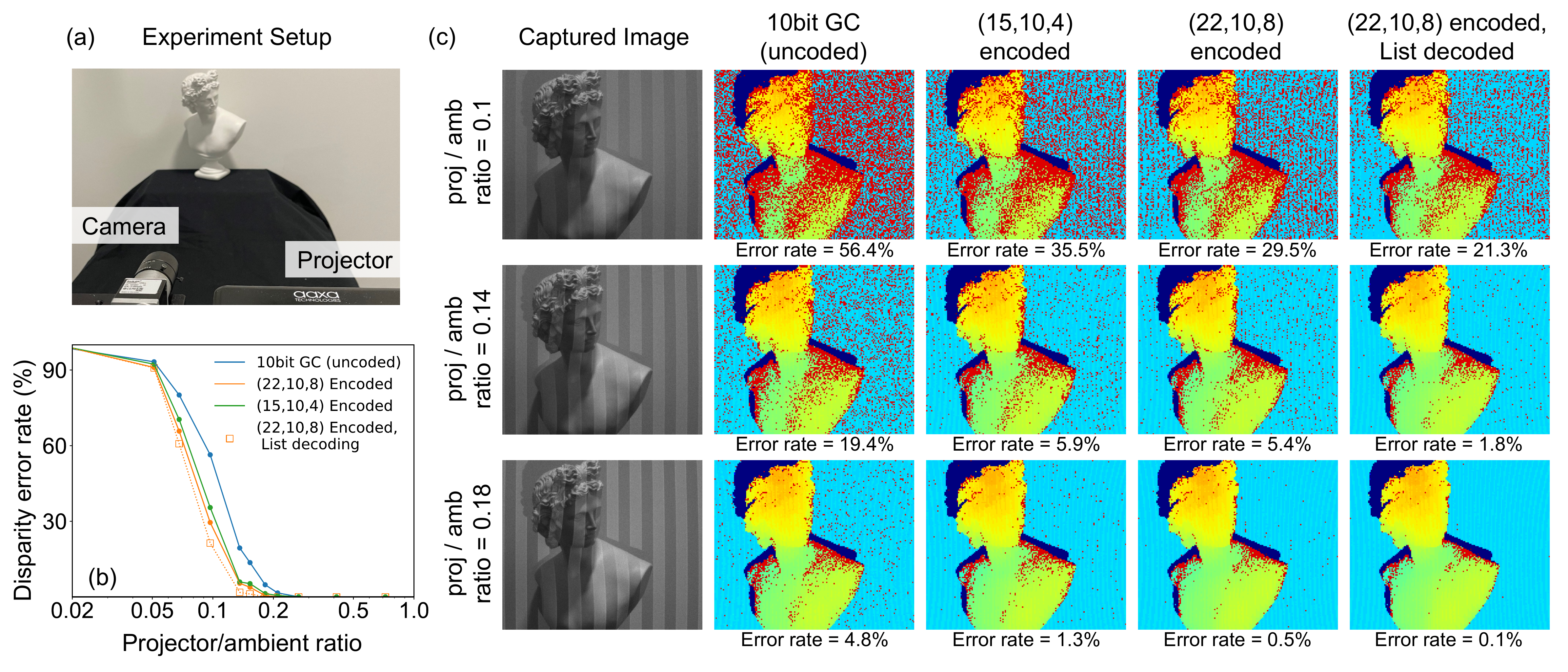}
\caption{Real world demonstration of ECC SL. (a)Experimental setup, with a camera, and a projector. (b) Quantitative comparison on SL performance (real-world data), with and without ECC, under different ambient light intensities. (c) Qualitative comparisons. All error pixels are marked out by red color. ECC SL has significantly lower error rate compared to the conventional SL.}
%\jian{is there improvement on new confidence map?}
\label{fig:ecc_real_world}
\end{figure*}

% We also evaluate the benefit of keeping more candidates in list decoding. As shown in Fig.~\ref{fig:decode_quant}, by keeping $7$ candidates instead of $3$ candidates in the correspondence search stage, error rate is further reduced. However, this benefit vanishes with increasing number of candidates, while computational cost increases linearly. Therefore, we regard keeping $3$ candidates in correspondence search as a good design choice.

We also show qualitative comparisons in Fig.~\ref{fig:soft_qual}. Soft-decision decoding is used to generate the results shown in Fig.~\ref{fig:soft_qual} (a). We mark out all erroneous pixels with red color for better visualization. ECC SL provides much cleaner disparity estimations compared to conventional SL under different ambient lighting conditions. In Fig.~\ref{fig:soft_qual} (b), we compare different decoding methods qualitatively. List decoding reduces the error rate by more than $2\times$ from the soft-decision decoded results. Also, it preserves the scene edges faithfully. The confidence-based median filter achieves a slightly lower error rate. Most errors are located at the depth discontinuities due to the comparatively large window size.

% Under Sunlight, we should use it. No sunlight and only readout noise, we shouldn't use it.
% Quantization noise (todo)

% N-ary codes can be used in SL system that can do more gray levels than 2 to decrease the number of total frames.
% Non-binary BCH codes (but I didn't find an example)
% 3-ary: Ternary Golay code, base 3, (12,6,d=6), which can differentiate $3^6=729$ columns. (11,5,d=6) can differentiate $3^5=243$ columns. Interestingly, under readout noise, it is not obviously worse than uncoded system, which is different from the binary case.
% 8-ary: Reed-Solomon (7,3,2) which can do $8^3=512$ columns. Readout noise is even better that uncoded system. 
% 16-ary: Reed-Solomon (15,3,6) can do $16^3=4096$ columns, which is more than enough. We can do shortened version, (15-1,3-1,6) which can do $16^2=256$ columns.

\subsection{Real-world experiments}
\label{sec:ecc_real_world}
% Leave this part to Jian.
% \jian{ Fig. 1;  Fig.2, golay+gray vs golay; Fig.3 adjust gain and expT, not work because of quantization noise;}\zh{done}

We build a hardware prototype to demonstrate ECC SL in real-world environments. As shown in Fig.~\ref{fig:ecc_real_world} (a), the experimental setup consists of one camera (Basler acA2040-120uc) and one DMD projector (Aaxa P300 Pico Projector), which is the same as a conventional SL system. We use a white lamp as the ambient light source and change the projector light intensity to tune the projector/ambient light ratios. We evaluate on the original 10bit GC, and two ECC codes, $(22,10,8)$ and $(15,10,4)$. We estimate a reference disparity map under strong projector light conditions. The reference disparity map estimations with three different codes have discrepancies smaller than $1$. Therefore, in the following experiments, we identify any pixel with a difference between estimated and reference disparities larger than $2$ as an error. Quantitative comparisons are shown in Fig.~\ref{fig:ecc_real_world} (b). It follows the same trend as that in the simulated environments. The maximum performance gap between ECC SL and conventional SL is $\sim 2\times$. We also show captured images (with original GC) and qualitative comparisons in Fig.~\ref{fig:ecc_real_world} (c). Similar to that in Fig.~\ref{fig:soft_qual}, we mark out all erroneous pixels with red color for better visualization. It is evident that ECC SL outperforms conventional SL significantly, while list decoding further improves the estimations.

We also examined the influence of code spatial arrangement and sensor readout noise in real-world environments. For detailed discussions, please refer to the supplementary material.

\section{SL with error detection codes}
% Jian's version, don't delete
\subsection{Principle of error detection codes (EDC)}
\label{sec: edc_principle}
We can further apply the EDC in SL systems. One family of codes, named ``cyclic redundancy codes'' (CRC), is specifically designed for error detection. They are short, and the encoding/error detection processes are more flexible compared to other ECC families.
%\jian{a little bit chaotic}\zh{done} 
In EDC, the received signal is first binarized and then goes through a generalized ``parity check'' process. If the received string does not fall in the code string codebook $C$, an error must exist.

For an ECC code, self-correction is guaranteed to succeed with less than $d_{min}/2$ number of errors. However, error detection is guaranteed to succeed with less than $d_{min}-1$ number of errors. In other words, when ECC fails, error detection still has a high probability of detecting that an error occurs (although the error can't be self-corrected). Then the receiver can request a re-sending from the sender for double-checking. An ECC that encodes $k$-bit strings into $n$-bit strings can detect at most $n-k-1$ errors, and the probability of getting an \textbf{undetectable} error scales as $2^{-(n-k)}$ in the best case~\cite{ecc_textbook}.

% Similar to that in , the detection process is achieved either with a ``hard-decision detection'' or a ``soft-decision detection'' process. For the hard-decision detection, the received signal is first binarized and then go through a ``parity-check'' process to determine whether it is within the code codebook or not. 

% by a generalized ``parity check'' on the received string~\cite{}. If the received string does not fall in the code string codebook $C$, an error is detected. One family of codes, named ``cyclic redundancy codes'' (CRC) are specifically designed for error detection and is widely applied in storage systems~\cite{}. They are short and the encoding/ error detection processes are more flexible compared to other ECC families.

% Various kinds of error detection codes. Read wikipedia and summarize them here.

% We choose CRC code, the probablity of false positive is quite low.
% $\eta = \frac{2^n - 2^k}{2^n} = 1 - 2^{-(n-k)}$. If $n=20, k=10$, $\eta = 0.999$ which means $0.001$ false positive happens.
% If $n=14, k=10$, $\eta = 0.9375$ which means $0.0625$ false positive happens. 

\subsection{How to use EDC in SL?}
EDC can be used to output an error map. If a camera pixel's received bits together do not pass the error checking, it means it has an error. Compared to the traditional contrast-based method where a contrast map is computed and then a threshold is manually set (a tradeoff between false negative and false positive) to check whether all bits are from measurements with good contrast, the error map based on EDC does not have a parameter to set, generally more accurate and has even lower computation. Since one method measures the individual bit's quality and the other measures the entire bit string's correctness, they could also be combined.

EDC can also be applied to an adaptive SL system~\cite{gupta2011structured, xu2009adaptive}, where an iterative process is employed to refine the reconstruction. This is necessary when ECC self-correction fails. 
% One example is, when part of the scene contains low albedo objects (e.g. black shelves). As discussed in Sec.~\ref{sec:simu_results}, in this scenario, self-correction capability in ECC is not beneficial. However, even under this . 
One example is when strong global illumination exists in the scene. Since codes at different spatial locations are mixed during inter-reflection and scattering, there are usually ``bursts'' of errors in the received signal instead of random errors in the strong ambient light scenario discussed in Sec.~\ref{sec:ecc}. The received signals are dissimilar to any codes in the codebook, and this leads to failure in error self-correction. However, such errors can be reliably detected.

% \jian{and the one before Mohit's adaptive}\zh{done}
Xu and Aliaga \cite{xu2009adaptive} and Gupta et al. \cite{gupta2011structured} proposed adaptive SL systems to handle global illumination. If a camera pixel's received projector bits are correct, the corresponded projector pixel will not illuminate in the next round. Since the amount of on-pixels decreases in each iteration, global illumination is decreased and the algorithm can converge.   
Gupta's method is based on four sets of patterns in each iteration, each with 10 bit planes. Disparity estimations are conducted with the 40 patterns, and pixels with inconsistent estimation results from the four codes are identified as erroneous. However, this is a great burden on the hardware system since $4$ times the amount of patterns is required for one refinement iteration. Here we demonstrate using EDC for the adaptive SL, while 
each iteration requires $1.4 - 2$ times the amount of original patterns. We enumerate several design choices on the EDC to be used.
\begin{itemize}
    \item Shorter EDC is preferred for smaller amount of frames in each iteration.
    \item EDC with large $d_{min}$ has higher error detection capability and is preferred, as explained in Sec.~\ref{sec: edc_principle}, 
    % \jian{Zhanghao explain why here}\zh{formula}
    \item  A spatially high-frequency pattern is better for SL with global illumination, as demonstrated in \cite{gupta2011structured}. 
    \item The EDC is better to be ``systematic''. That is, the message bits do not change after encoding. With this property, we could directly exploit the pattern design in the literature.
    % The appended EDC bits should also be as high frequency as possible.  
    % \jian{By this, we could exploit the pattern design in the literature.} 
\end{itemize}

% After one round of pattern projections, part of the pixels are classified as erroneous, and are illuminated again in the next round, while other pixels are turned off. This process effectively removes large portions of global illumination in the refinement stage~\cite{}, and is thus robust to challenging scenes with heavy inter-reflections or sub-surface scatterings. Previous research~\cite{} projects multiple sets of patterns, and reconstruct the scene geometry separately. Pixels that are regarded as erroneous pixels. This process is time-consuming, and \textcolor{red}{XXX}. We list several constraints in selecting an appropriate error detection code:

% Binary code + CRC, Gray code + CRC, Mohit's code \cite{gupta2011structured} + CRC (this one)
% \textbf{Extension to N-ary code} \cite{zhou2000non} Zhanghao: I didn't read this paper. So read it and write something here. You can also search ``non binary code, CRC'' to look for other papers

With these design choices, we examined different combinations between high-frequency patterns and ECC/EDC codes, and we found it non-trivial to meet all the preferred properties. It turns out that XOR02 code from~\cite{gupta2011structured} $+$ CRC5 code from the CRC family is a good candidate (we also show several unfavorable combinations in the supplementary material). As shown in Fig.~\ref{fig:edc_high_freq} lower row. It appends $5$ extra frames behind the $10$ frames XOR02 patterns, achieving $d_{min} = 4$. High-frequency property is also maintained. %\jian{include two failure cases in supp}\zh{done} \zh{Leave to Jian: there is no code for max-min width.}

% With these principles, we searched code + CRC of various kinds, code + ECC, and found that (1) for high freq., there exists one which is shown in Fig. (2) 

\begin{figure}[!t]
\centering 
\includegraphics[width=0.9\columnwidth]{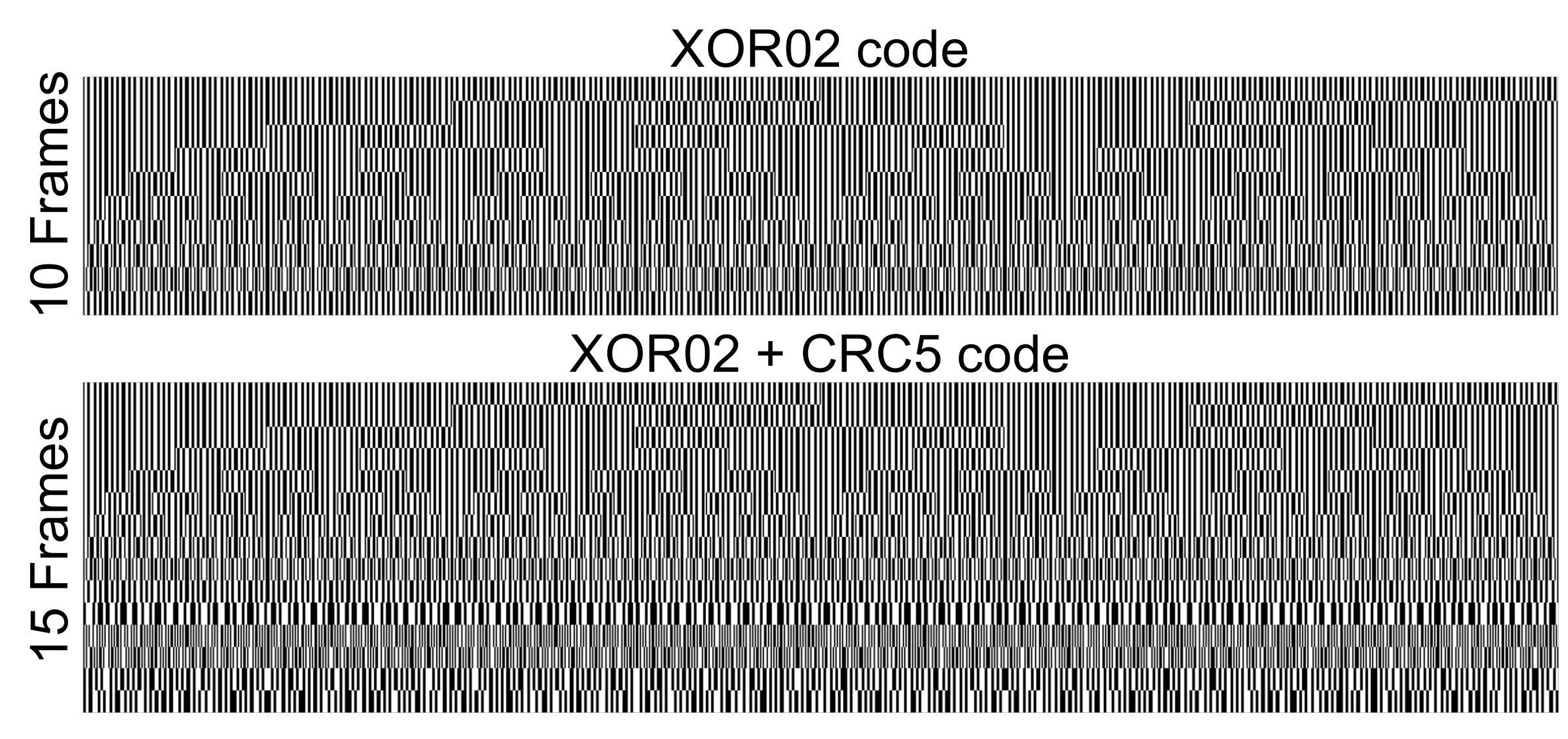}
\caption{Error detection code (CRC) for SL with global illumination. (a) High frequency XOR02 code proposed in ~\cite{gupta2011structured}. (b), XOR02 code with 5bits CRC code appended, for error detection and adaptive SL.}
%\jian{which CRC5}
\label{fig:edc_high_freq}
\end{figure}

\begin{figure}[!t]
\centering 
\includegraphics[width=0.6\columnwidth]{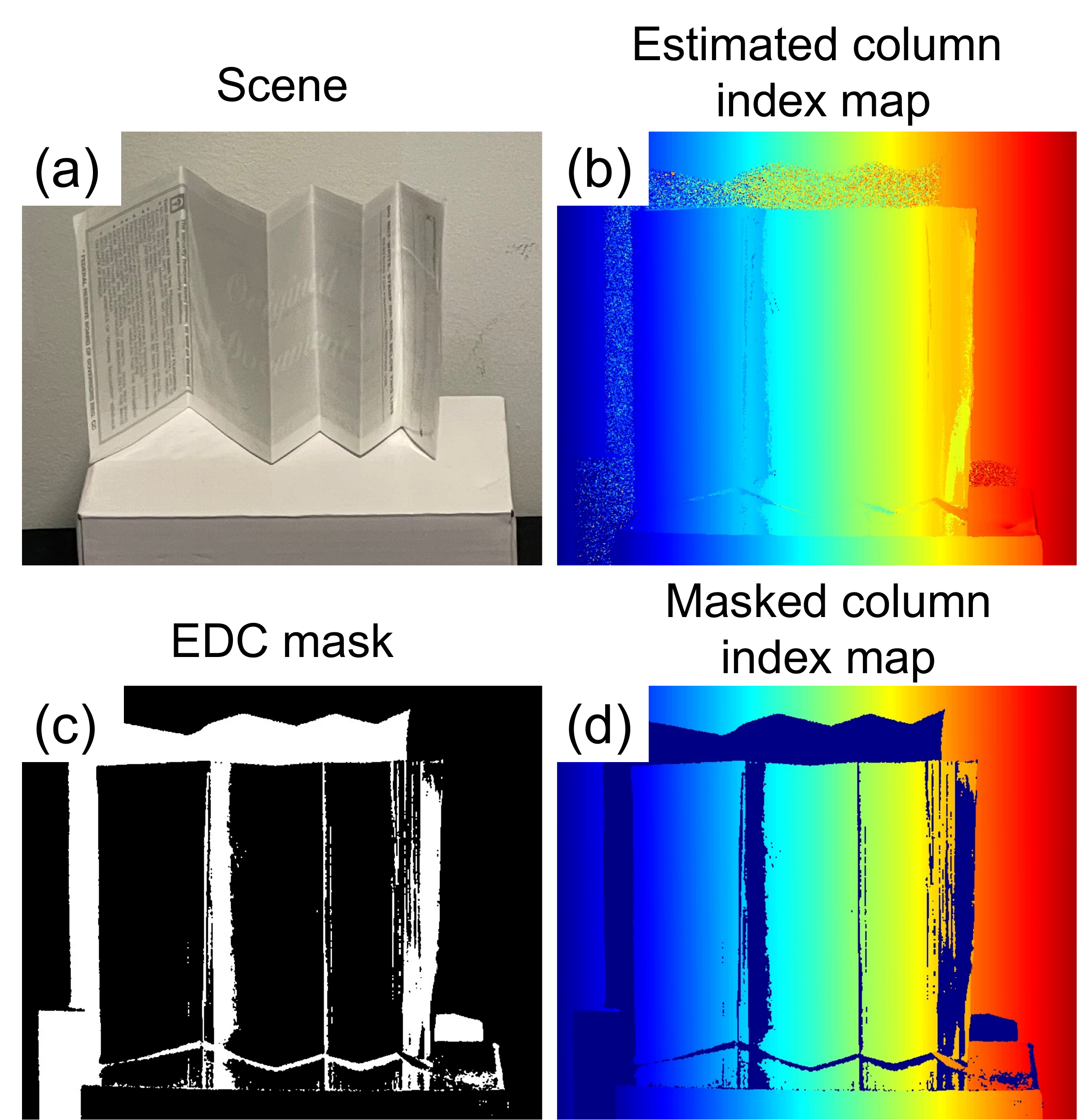}
\caption{EDC error detection with inter-reflection. (a) Scene with V-grooves as source of inter-reflection. (b) Estimated column index map. (c) EDC mask (white color denotes \textbf{incorrect} estimations). Notice that the regions with shadows and inter-reflections are detected correctly. (d) Masked column index map.}
\label{fig:edc_error_map}
\end{figure}

\begin{figure*}[t]
\centering 
\includegraphics[width=0.7\textwidth]{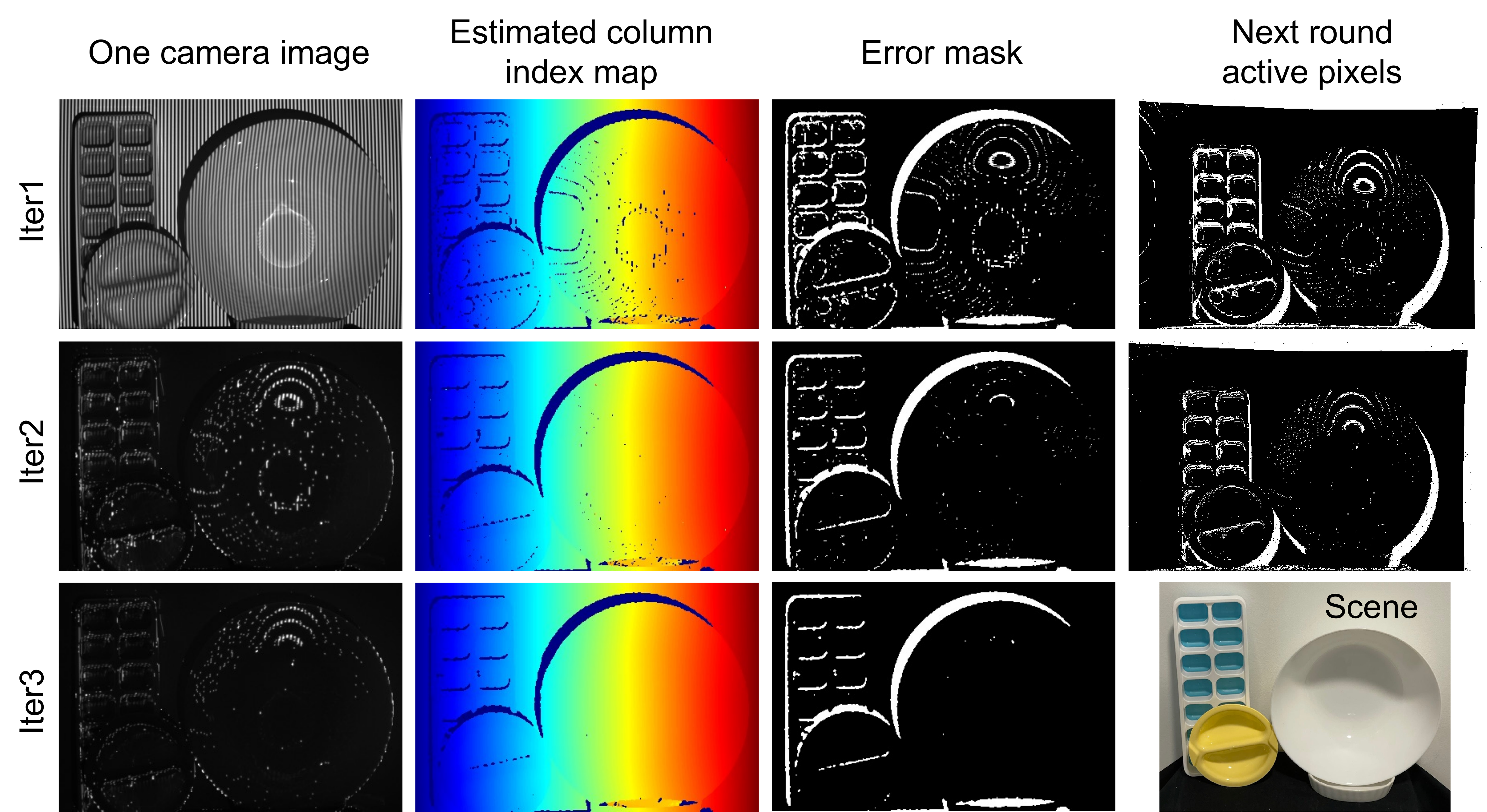}
\caption{Real-world demonstration on EDC based adaptive SL. The whole process converges in three iterations ($45$ frames in total), and most artifacts in disparity estimation is removed.} % \jian{why Mohit didn't use contrast map. Byeonjoo's SL work by EDC}}
\label{fig:edc_iter}
\end{figure*}

\subsection{Real-world experiments}
We first show an EDC generated error detection map in a scene with strong V-groove inter-reflections (Fig.~\ref{fig:edc_error_map}). Since the designed high-frequency pattern handles this simple geometry well, to validate the effectiveness of EDC, we combine the CRC5 EDC with the original gray code in this proof-of-concept experiment. As expected, directly estimating column correspondence in this scene results in large artifacts close to the hinges, as shown in Fig.~\ref{fig:edc_error_map} (b). However, with the $5$bits redundant EDC, these errors can be detected, as shown in Fig.~\ref{fig:edc_error_map} (c), (d). 

% We first show an error map by EDC. EDC can be appended to any code and here we use gray code + CRC5 as an example. It is shown in Fi.g 

Starting from the EDC error map, we can conduct the iterative process to refine estimation. We demonstrate this full process with designed XOR02+CRC5 pattern in a very challenging scene. The scene consists of plastic and ceramic containers. Inter-reflection, sub-surface scattering, and other complicated light paths co-exist. As shown in Fig.~\ref{fig:edc_iter}, within three iterations, the adaptive error-detection-projection-measurement process converges. Most artifacts due to global illumination and scattering are removed. The whole acquisition process contains $45$ frames, which is roughly the same amount of frames required in one iteration of the previous adaptive SL system~\cite{gupta2011structured}.
% with We show iterative SL results in Fig. .

% end of Jian's version

\section{SL with light source encoding}
Apart from the error correction and detection codes, we introduce another type of redundancy codes to avoid interference. We named it ``light source encoding'' since it encodes each light source state (on/off) individually, instead of encoding the whole temporal sequence as that in ECC/EDC. In this section, we first explain the proposed method, and then show two examples.

\begin{figure*}[!t]
\centering 
\includegraphics[width=1.0\textwidth]{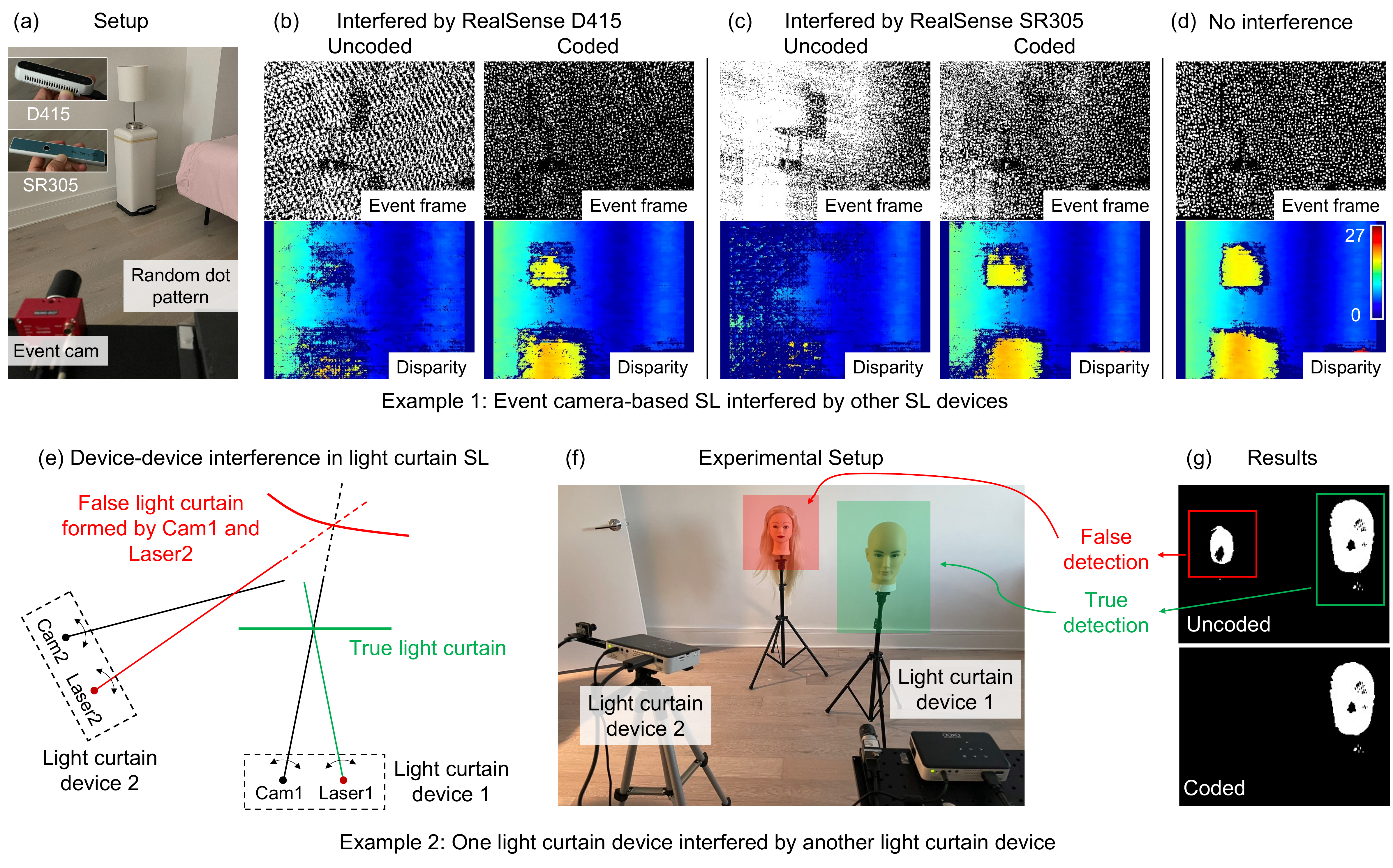}
\caption{We use light source encoding to relieve the interference by other active illumination systems.}
\label{fig:source_encoding}
\end{figure*}

\subsection{Principle of light source encoding}
SL systems often suffer from interference from other active illumination systems or dynamic ambient light reflections (due to either dynamic scene or dynamic ambient light source). Although the camera and the projector within one SL system are temporally synchronized, the interference can't be totally removed. The camera and the projector can also be ``spectrally synchronized'' by adding a light source-matched narrow wavelength bandpass filter to the camera. However, this reduces the light efficiency of the system and requires different hardware designs. The active stereo system solves this problem with an additional camera, which is effectively a redundancy on the hardware side.

We propose to solve this problem by encoding the projector's temporal illumination sequence into a more distinguishable form with redundancy codes. We want each SL system to use a device-specific code. Specifically, we transform each `1' (light on) in the projection sequence into a device-specific bit string (e.g., `1101' for one device and `1010' for another device). This process is similar to the code-division-multiplexing (CDM) in communication systems, which enables different users to share the same cellular network channel. The longer the bit string, the more devices are compatible to share the same space, and the robustness against external interference is higher. Generally, this method can be applied in any SL system, including the ECC/EDC SL discussed above. However, the code length is heavily constrained by the speed of the imaging system. Therefore, it is more beneficial to apply such technique in specific SL systems that are inherently high frame rate, e.g. event camera-based SL \cite{matsuda2015mc3d} and line sensor-based SL \cite{wang2016dual}. Here we demonstrate such applications in real-world prototypes.

\subsection{Example 1: event-camera based SL}
The system consists of an event camera and a structured light source that projects a random dot pattern. When the projection is set to an ``On-off'' mode, events are triggered in the camera due to temporal intensity change. Since the events are only triggered at the pattern dot positions, the amount of acquired data is largely reduced and enables a high frame rate and more efficient post-processing. Block matching algorithm is performed to estimate the disparity map from the ``event frame'' obtained from an on-off period, as shown in Fig.~\ref{fig:source_encoding} (d). However, when an external dynamic light source (e.g., an Intel Realsense device) is close-by, the event camera is also triggered by the illumination change from it, and this interference degrades the performance dramatically (as shown in Fig.~\ref{fig:source_encoding} (b),(c) left part). To solve this problem, we encode each on-off period to be `10100010', instead of `10' (`1' stands for on, `0' stands for off), with $4\times$ more frames. We conduct a matched filter on the event frames to remove the events coming from interference (which does not match the designed time sequence). As shown in the upper right of Fig.~\ref{fig:source_encoding} (b),(c), the filtered event frame becomes much cleaner, and can produce reasonable disparity estimations (lower right of Fig.~\ref{fig:source_encoding} (b),(c)).
% don't delete. If the system is moving, there could also be events coming from the scene, which is not by the projector.
% The system is composed of an event camera and a random dot pattern (see Fig. \ref{fig:source_encoding} (a.1)). The pattern `being on' generates 
%positive 
% events in the camera. By collecting events within a short duration, an event frame is formed. We can then use block matching algorithm to get the disparity map, as shown in (a.4). When an Intel RealSense is operated nearby, it will interfere the SL system dramatically. We can see RealSense 415's pattern (a.2 left above) and RealSense SR305-caused huge amount of false positive events (a.3 left above) in the event frame which cause severe degradation in the calculated disparity maps ((a.2) and (a.3) left below). When adding the code to the light source, here we use `1101', we get four event frames to calculate one disparity map. The filtered event frames are obtained by checking whether the events at each pixel follow the `1101' pattern and are shown in (a.2) and (a.3) right above. They are more cleaned and the disparity maps calculated from them have much less artifacts, as shown in (a.2) and (a.3) right below.

\subsection{Example 2: line sensor based SL - light curtain}
A light curtain (LC) device \cite{wang2018programmable} is a special SL system. It consists of a line sensor and a line laser. The light plane and the viewing plane intersect at a line in the 3d space, which forms a triangular geometry as in a conventional SL system. Any object located at that line will be detected by the line sensor. By spatially scanning the viewing directions of the line sensor and the line laser, the intersection line scans a curved plane defined by the user, named the ``light curtain''. The device is able to concentrate optical power to get high SNR and can be applied in obstacle detection~\cite{bartels2019agile,ancha2020active}. When two such LC devices are close to each other, interference causes false detections. One example is shown in Fig.~\ref{fig:source_encoding} (e), a false detection curtain is formed due to interference between the camera from the first LC device and the laser in the second LC device. We emulate this effect in a real-world environment with two camera-projector pairs, as shown in Fig.~\ref{fig:source_encoding} (f). The scanning mechanism is achieved by only lightening one projector column and capturing data at one camera column in each frame. By aggregating the frames, a light curtain image is formed, as shown in Fig.~\ref{fig:source_encoding} (g), which highlights the detected objects. When the first LC device is scanning the frontal-parallel plane close to the male mannequin, a false light curtain is formed between the light source in LC2 and the camera in LC1. This results in false detections located around the female mannequin, as shown in the upper part of Fig.~\ref{fig:source_encoding} (g). We encode the two light sources with different codes `1100' and `0101', respectively. After acquiring all the frames ($4\times$ compared to the original LC device), only pixels with the correct intensity sequence are kept, and the false detection is easily removed, as shown in the lower part of Fig.~\ref{fig:source_encoding} (g).

% By rotating the line sensor and line laser, a bunch of lines, forming a curved plane which is named as light curtain, are scanned. The device has several good properties and can be used to avoid obstacles \cite{bartels2019agile,ancha2020active}. However, when two LC devices are put together, they can easily interfere each other. As shown in Fig. \ref{fig:source_encoding} (b.1), a false positive curtain could be formed by a camera from one LC device and a laser from another LC device. In the setup shown in (b.2), LC device 1 is scanning a frontal-parallel plane at the the depth of the male mannequin (due to the difficulty of building the hardware, we use a camera and a projector to simulate an LC device). Because of the second LC device, a false positive curtain is generated and the female mannequin at another depth is falsely detected (shown in (b.3) above). We propose to encode the lasers of the two devices by `1100' and `0101', respectively. After getting four line sensor images, we normalize them and check whether the detections match the code. In this way, we easily remove the false positive detection, as shown in (b.3) below.

\section{Conclusion}
In this paper, we propose a general redundancy coding scheme for SL systems to mitigate various types of degradation, including noise, global illumination, and interference. We demonstrate error correction, error detection, and light source encoding based on redundancy codes. Both simulations and real-world experiments demonstrate a significant ``coding gain'' in performance. The proposed technique has the potential to be implemented as a supplemental plug-in to previously developed SL techniques, with a minimum amount of hardware modifications required. This leads to improved robustness and accuracy in applications across disciplines, including augmented reality, industrial inspection, and remote sensing. Moreover, the proposed redundancy coding could also augment other similar architectures like time-of-flight devices or a broader category of active illumination imaging systems.

% \zh{Leave to Jian} Moreover, these methods can be extended to direct ToF directly, since TOF is also susceptible to errors under ambient illumination, global illumination and multi-path interference.

% Any acknowledgments to only be included in camera ready

\section*{Acknowledgments}
The authors would like to thank Prof. Shree K. Nayar %and Mohit Gupta 
for helpful discussions, thank Gurunandan Krishnan for general support and discussion of event camera-based structured light, thank Joe Bartels, Siddharth Ancha and Gaurav Pathak for the light curtain part, and thank Wenzheng Chen for the discussion of optimal code design.
%Prof. Mohit Gupta (add or not)

% \newpage
\appendices
% \section{Supplementary}
\section{Code lookup table}
\begin{figure}[!t]
\centering 
\includegraphics[width=1.0\columnwidth]{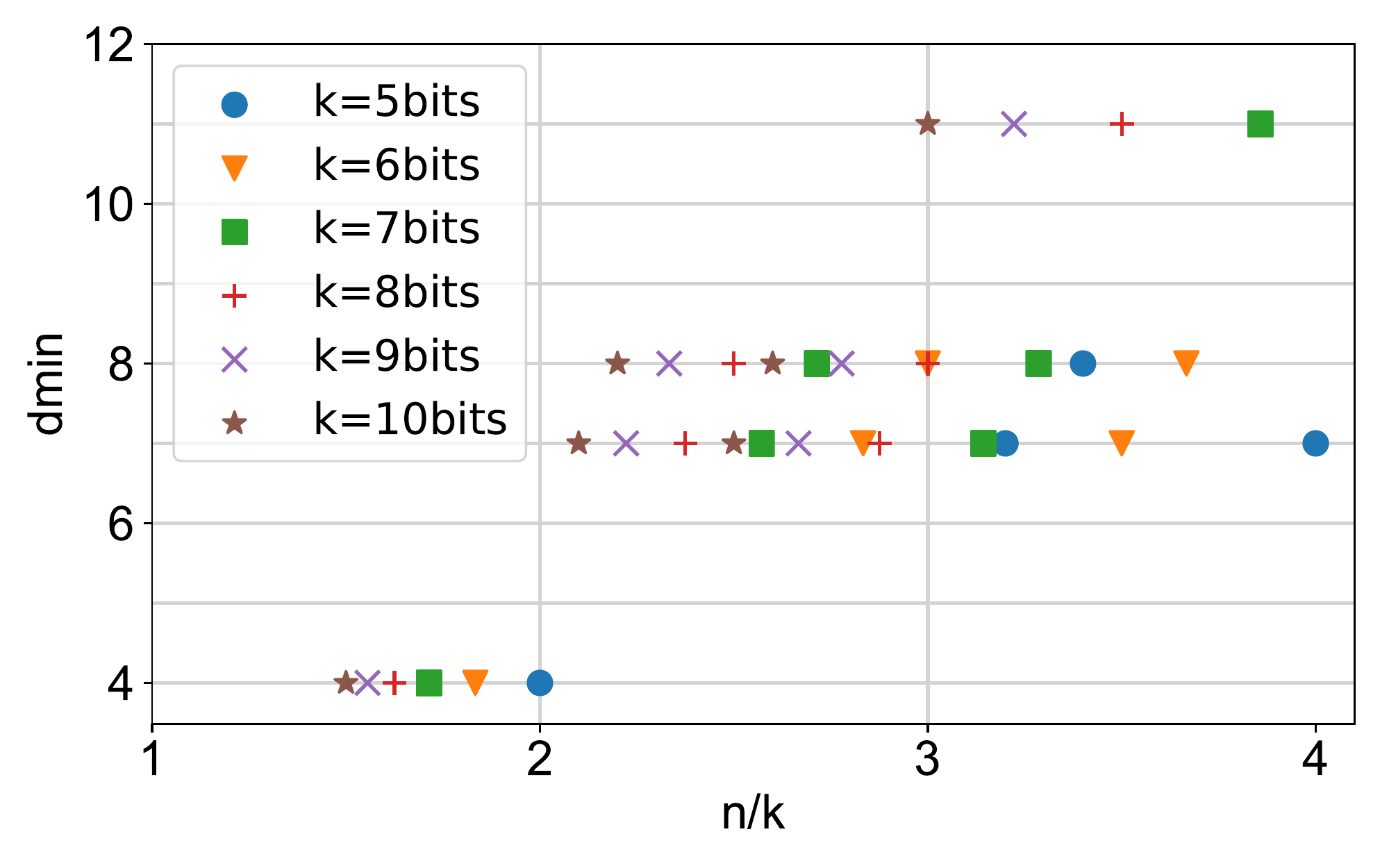}
\caption{Code lookup table for different encoded bit length $n$ and original bit length $k$.}
\label{fig:code_lut}
\end{figure}

\begin{figure}[!t]
\centering 
\includegraphics[width=1.0\columnwidth]{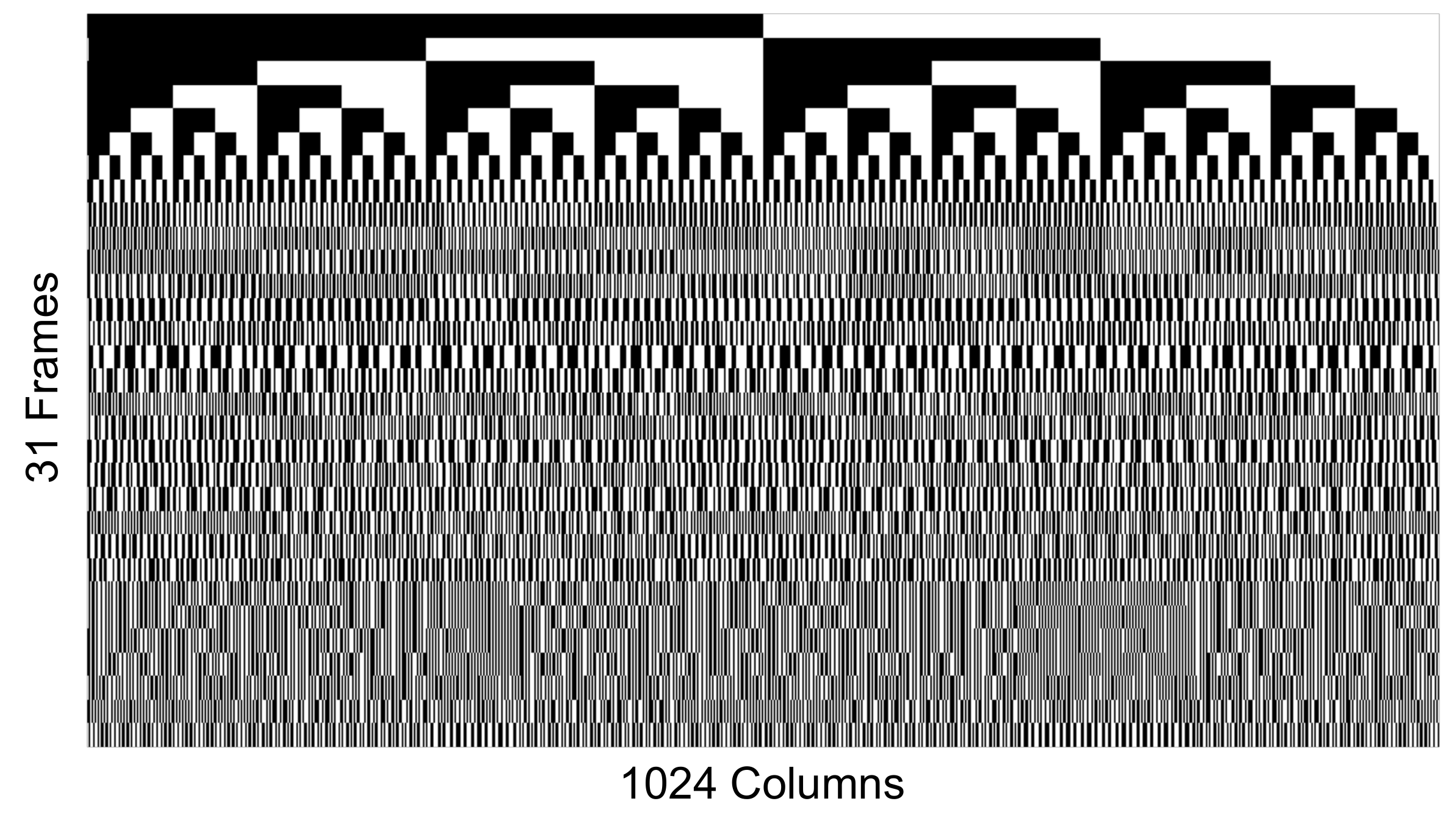}
\caption{A (31,10,12) pattern generated from random search.}
\label{fig:random_search}
\end{figure}

Here we show the code lookup table (Fig.~\ref{fig:code_lut}) discussed in main text, Sec. 3.2.3. For each data bit string length $k$ (corresponding to the resolution of the SL system), and total bit string length $n$ (corresponding to the frame rate of the SL system), we select the best performance code from these families. Since the codes are conceived following certain algebraic rules, ECC only exists with special $n, k$ values. Fortunately, all the code families we examined (also most high-performance code families) are ``systematic'' codes (or can be transformed to equivalent systematic codes), where the encoders only append error correction codes after the data bit-string and do not change the data bit string. This allows straightforward bit string truncation. By setting the first $L$ bits in the data bit-string to be $0$ and removing them in the encoded bit string, we turn an $(n,k)$ ECC into an $(n-L, k-L)$ ECC, with at least same $d_{min}$. We also contain a code generated from a high-dimensional Poisson disk sampling algorithm that does not belong to any of these designed code families but has comparative $d_{min}$, as shown in Fig.~\ref{fig:random_search}.

\iffalse
\section{Approximate nearest neighbor search in decoding}
Since most computations in the decoding process occur in correspondence search between received signals and codes in the codebook, an approximate nearest neighbor (ANN) search algorithm can be utilized to increase processing speed by $\sim 5\times$. It also supports outputting multiple candidates in the search process, which is useful in list decoding. We use the PyFLANN package~\cite{pyflann}, and use $10$ iterations, $128$ checks as input parameters. We observe a speed up from 40 seconds to 5 seconds ($\sim 8\times$) for $600\times700$ resolution disparity maps decoding on a single Intel Xeon CPU core. We show quantitative comparisons on accuracy in Fig.~\ref{fig:ann}. It can be seen that the error rate is almost the same with and without the approximation.
\fi
\section{Readout noise and quantization error}
As discussed in the main text Sec. 3.2.5, when sensor readout noise and quantization error dominates, ECC SL has similar or even worse performance compared to conventional SL. This is majorly due to the fact that per-frame SNR scales with per-frame exposure time $t_{exp}$ linearly in this regime, while it scales with $\sqrt{t_{exp}}$ when shot noise dominates. 

\begin{figure}[!t]
\centering 
\includegraphics[width=0.8\columnwidth]{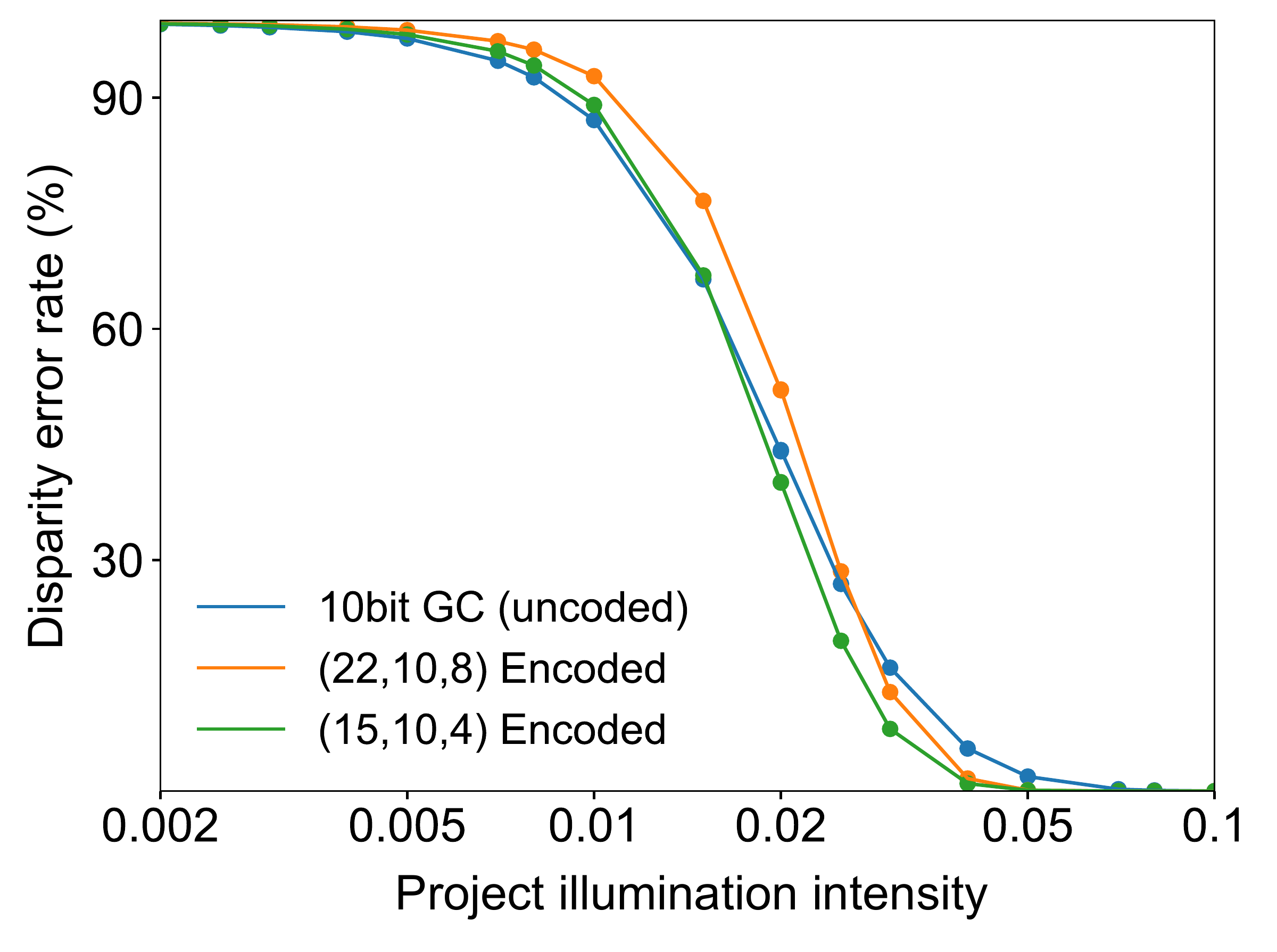}
\caption{Quantitative comparisons when readout noise and quantization error dominates.}
\label{fig:readout_noise}
\end{figure}

\begin{figure}[!t]
\centering 
\includegraphics[width=1.0\columnwidth]{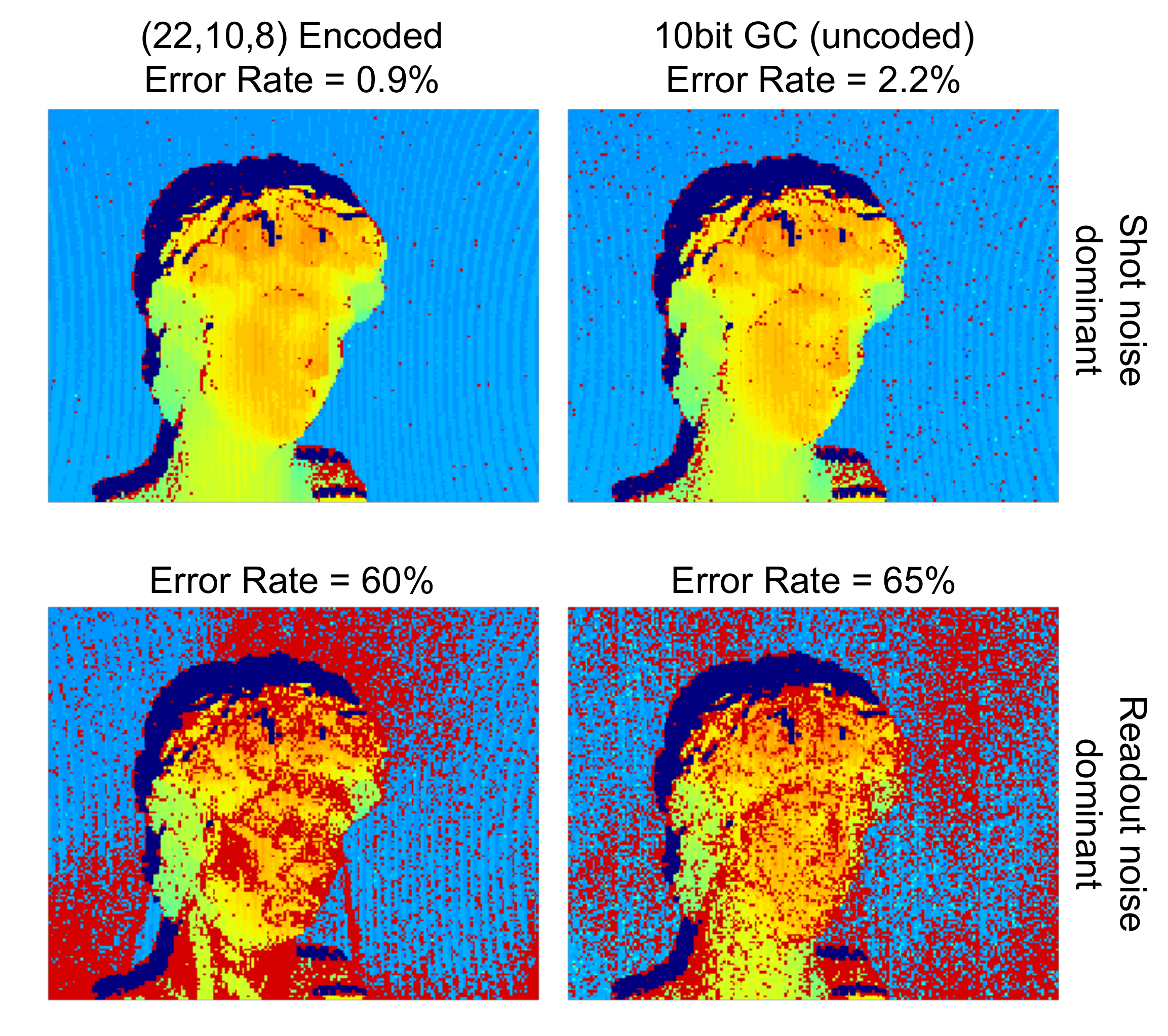}
\caption{Readout noise and quantization error dominant scenario in real-world experiments.}
\label{fig:readout_noise_real}
\end{figure}

Here we show this effect in both simulated and real-world environments. We use the same FlyingThings3D synthetic dataset~\cite{MIFDB16} for simulation. We set the ambient illumination to be zero and set the signal level of projector illumination to be smaller than $0.1$. We use a readout noise $\sigma_g = 0.004$ and a $12$bit quantization. In Fig.~\ref{fig:readout_noise}, we show quantitative comparison on disparity error rate. In Fig.~\ref{fig:readout_noise_real}, we further confirm this effect in real-world experiment. 

\section{Code spatial arrangement}
\begin{figure}[!t]
\centering 
\includegraphics[width=0.7\columnwidth]{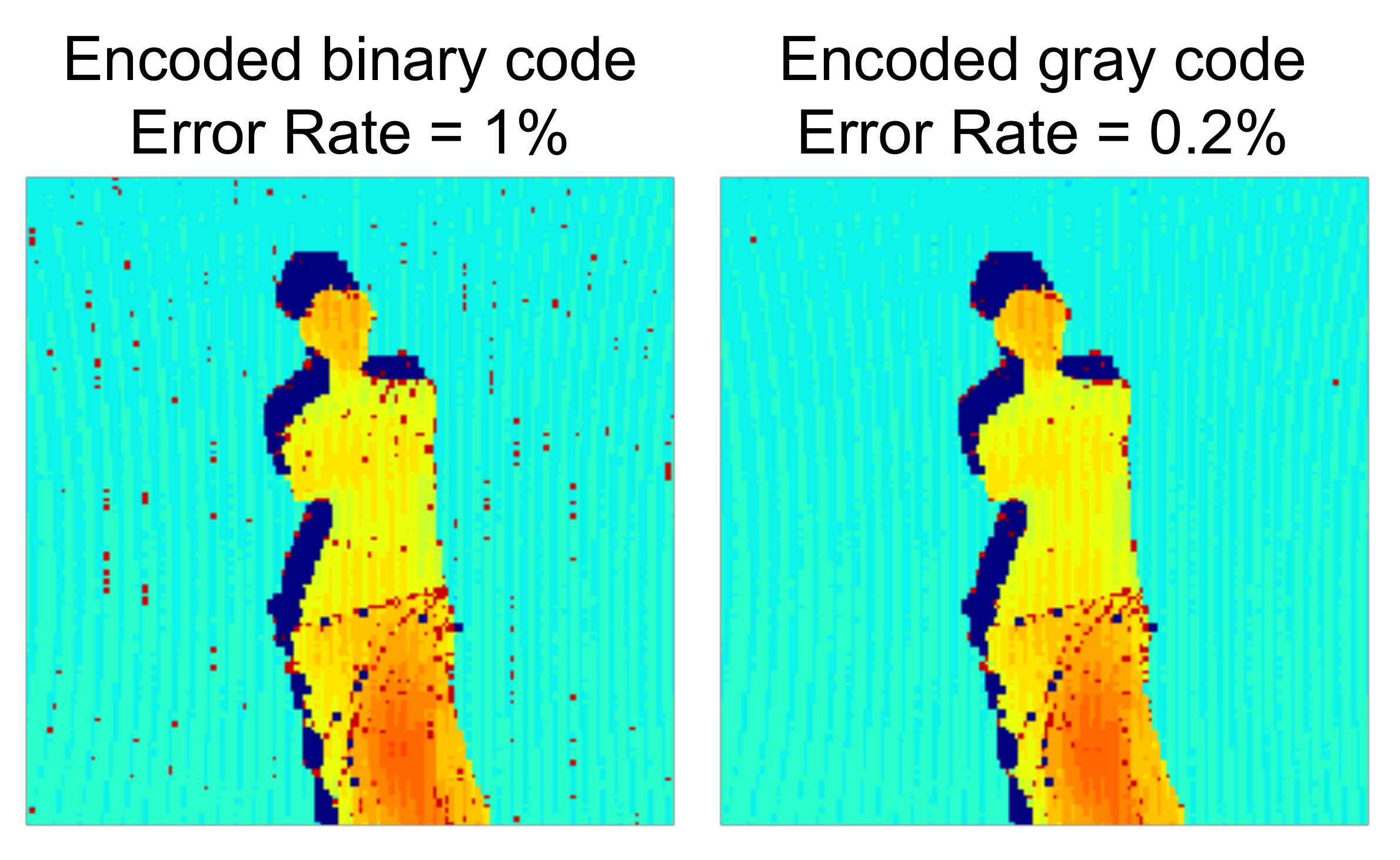}
\caption{Effect of code arrangement in real-world experiments. ECC encoded gray code has smallest neighborhood difference, and thus has lower error rate compared to the encoded binary code.}
\label{fig:arrange_real_world}
\end{figure}

Similar to the conventional SL patterns, the arrangement of columns in ECC SL patterns also matters. Less `0-1' jumps in the pattern reduce artifacts in real-world experiments~\cite{geng2011structured}. Compared with binary code patterns, gray code guarantees Hamming distance $=1$ between adjacent columns. In ECC SL, since the minimum pair-wise distance in the bit-string codebook is $d_{min}$, this is the minimum possible difference between adjacent columns. Empirically, we found this constraint is satisfied when using gray code as the input to the two encoders used in our real-world experiments ($(22,10,8)$ and $(15,10,4)$). We show this effect in Fig.~\ref{fig:arrange_real_world}. We observe that the ECC encoded from gray code has less error compared to the other one encoded from binary code with the same encoder.

\section{ECC in N-ary gray-level SL}
\begin{figure}[!t]
\centering 
\includegraphics[width=1.0\columnwidth]{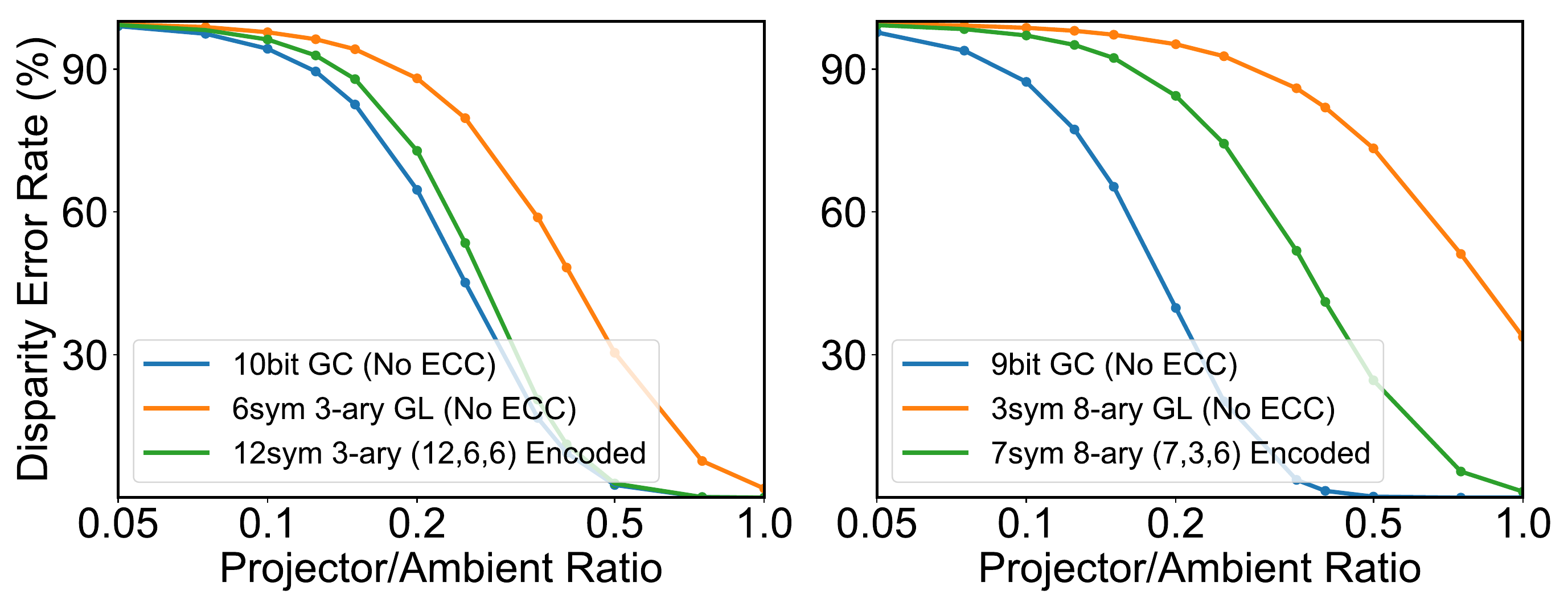}
\caption{Two examples of N-ary gray level SL, with and without ECC. Although ECC enhances the disparity estimation accuracy, N-ary SL systems are still less robust than binary SL systems with similar number of frames under strong ambient illumination.}
\label{fig:n-ary}
\end{figure}

Another type of temporally multiplexed SL system use N-ary gray-level patterns instead of binary patterns discussed above~\cite{n-ary_SL}. This reduces the number of frames needed to achieve the same spatial resolution. Similarly, in communication systems, N-ary ECC is designed for N-ary symbol transmissions. As an example, Reed-Solomon code~\cite{reed_solomon} is one of the most commonly used N-ary ECC (e.g., in QR codes). However, the robustness of N-ary SL is not as high as binary SL. The N-ary pattern reduces the number of frames by $\textrm{log}_2N$ times. However, the interval between projector light intensity levels becomes $N-1$ times smaller. Therefore, given a fixed total exposure time and projector light intensity, the effective SNR reduces $\frac{N-1}{\sqrt{\textrm{log}_2N}}$ times when shot noise and ambient light dominates. This factor increases with bigger $N$. We compare the quantitative performance of the binary SL and N-ary SL systems, with or without ECC, in Fig.~\ref{fig:n-ary}. We used 3-ary $(12,6,6)$ Golay code and 8-ary  $(7,3,6)$ Reed-Solomon codes as examples, which has spatial resolutions $729$ and $512$. Accordingly, we use $10$bits and $9$bits binary patterns for comparison. Although N-ary ECC patterns indeed reduce the error rate compared with the uncoded N-ary patterns, their robustness are similar or worse than uncoded binary patterns with similar spatial resolution. Therefore, previously proposed self-correction methods based on N-ary gray-level SL~\cite{li2021error, porras2017error} fail in this challenging situation.
 
\section{Unfavorable codes in EDC}
\begin{figure}[!t]
\centering 
\includegraphics[width=1.0\columnwidth]{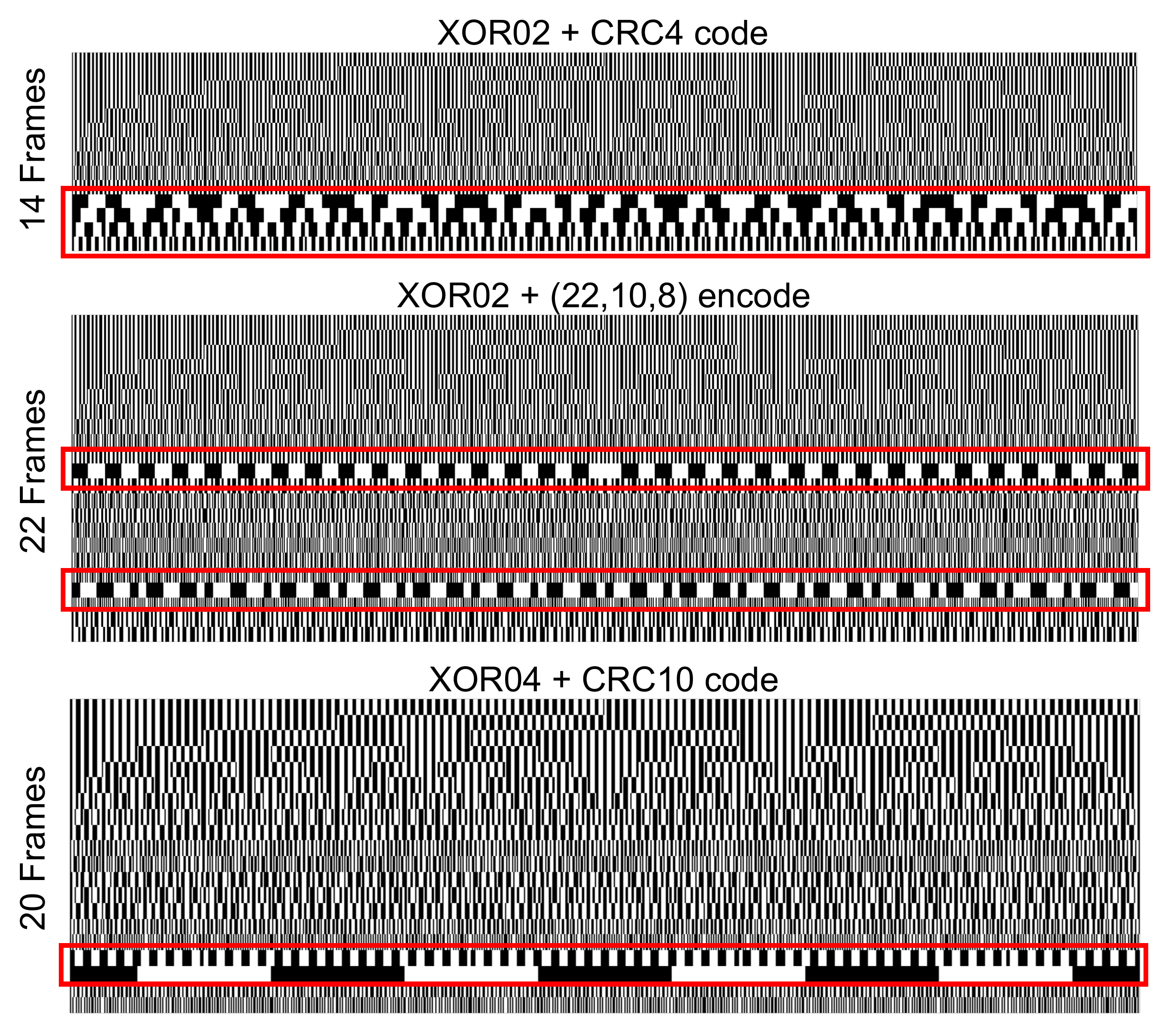}
\caption{Three unfavorable combinations of high-frequency codes proposed in ~\cite{gupta2011structured} and ECC/CRC encoding. Frames that violate high-frequency property are shown in red bounding boxes.}
\label{fig:edc_fail}
\end{figure}

Combinations of the XOR02 and XOR04 codes proposed in ~\cite{gupta2011structured} and ECC/EDC codes are not always successful. We show three such examples in Fig.~\ref{fig:edc_fail}. Although the data frames have high spatial frequency by design, the ECC/EDC frames might not maintain this favorable property.

\iffalse
\begin{figure}[!t]
\centering 
\includegraphics[width=0.8\columnwidth]{ann.pdf}
\caption{Quantitative comparisons with and without approximation in nearest neighbor search.}
\label{fig:ann}
\end{figure}
\fi

\bibliographystyle{IEEEtran}
\bibliography{ref}

\begin{thebibliography}{10}
\providecommand{\url}[1]{#1}
\csname url@samestyle\endcsname
\providecommand{\newblock}{\relax}
\providecommand{\bibinfo}[2]{#2}
\providecommand{\BIBentrySTDinterwordspacing}{\spaceskip=0pt\relax}
\providecommand{\BIBentryALTinterwordstretchfactor}{4}
\providecommand{\BIBentryALTinterwordspacing}{\spaceskip=\fontdimen2\font plus
\BIBentryALTinterwordstretchfactor\fontdimen3\font minus
  \fontdimen4\font\relax}
\providecommand{\BIBforeignlanguage}[2]{{%
\expandafter\ifx\csname l@#1\endcsname\relax
\typeout{** WARNING: IEEEtran.bst: No hyphenation pattern has been}%
\typeout{** loaded for the language `#1'. Using the pattern for}%
\typeout{** the default language instead.}%
\else
\language=\csname l@#1\endcsname
\fi
#2}}
\providecommand{\BIBdecl}{\relax}
\BIBdecl

\bibitem{gupta2011structured}
M.~Gupta, A.~Agrawal, A.~Veeraraghavan, and S.~G. Narasimhan, ``Structured
  light 3d scanning in the presence of global illumination,'' in
  \emph{Proceedings of the IEEE International Conference on Computer
  Vision}.\hskip 1em plus 0.5em minus 0.4em\relax IEEE, 2011, pp. 713--720.

\bibitem{gupta2013structured}
M.~Gupta, Q.~Yin, and S.~K. Nayar, ``Structured light in sunlight,'' in
  \emph{Proceedings of the IEEE International Conference on Computer Vision},
  2013, pp. 545--552.

\bibitem{maimone2012reducing}
A.~Maimone and H.~Fuchs, ``Reducing interference between multiple structured
  light depth sensors using motion,'' in \emph{2012 IEEE Virtual Reality
  Workshops}.\hskip 1em plus 0.5em minus 0.4em\relax IEEE, 2012, pp. 51--54.

\bibitem{cdma}
A.~J. Viterbi, \emph{CDMA: principles of spread spectrum communication}.\hskip
  1em plus 0.5em minus 0.4em\relax Addison Wesley Longman Publishing Co., Inc.,
  1995.

\bibitem{salvi2004pattern}
J.~Salvi, J.~Pages, and J.~Batlle, ``Pattern codification strategies in
  structured light systems,'' \emph{Pattern recognition}, vol.~37, no.~4, pp.
  827--849, 2004.

\bibitem{posdamer1982surface}
J.~L. Posdamer and M.~Altschuler, ``Surface measurement by space-encoded
  projected beam systems,'' \emph{Computer graphics and image processing},
  vol.~18, no.~1, pp. 1--17, 1982.

\bibitem{inokuchi1984range}
S.~Inokuchi, ``Range imaging system for 3-d object recognition,'' \emph{ICPR,
  1984}, pp. 806--808, 1984.

\bibitem{bergmann1995new}
D.~Bergmann, ``New approach for automatic surface reconstruction with coded
  light,'' in \emph{Remote sensing and Reconstruction for three-dimensional
  objects and scenes}, vol. 2572.\hskip 1em plus 0.5em minus 0.4em\relax SPIE,
  1995, pp. 2--9.

\bibitem{hall2001stripe}
O.~Hall-Holt and S.~Rusinkiewicz, ``Stripe boundary codes for real-time
  structured-light range scanning of moving objects,'' in \emph{Proceedings of
  IEEE International Conference on Computer Vision}, vol.~2.\hskip 1em plus
  0.5em minus 0.4em\relax IEEE, 2001, pp. 359--366.

\bibitem{morita1988reconstruction}
H.~Morita, K.~Yajima, and S.~Sakata, ``Reconstruction of surfaces of 3-d
  objects by m-array pattern projection method,'' in \emph{1988 Second
  International Conference on Computer Vision}.\hskip 1em plus 0.5em minus
  0.4em\relax IEEE Computer Society, 1988, pp. 468--469.

\bibitem{sarbolandi2015kinect}
H.~Sarbolandi, D.~Lefloch, and A.~Kolb, ``Kinect range sensing:
  Structured-light versus time-of-flight kinect,'' \emph{Computer vision and
  image understanding}, vol. 139, pp. 1--20, 2015.

\bibitem{maruyama1993range}
M.~Maruyama and S.~Abe, ``Range sensing by projecting multiple slits with
  random cuts,'' \emph{IEEE Transactions on Pattern Analysis and Machine
  Intelligence}, vol.~15, no.~6, pp. 647--651, 1993.

\bibitem{hung19933d}
D.~D. Hung, ``3d scene modelling by sinusoid encoded illumination,''
  \emph{Image and Vision Computing}, vol.~11, no.~5, pp. 251--256, 1993.

\bibitem{tajima19903}
J.~Tajima and M.~Iwakawa, ``3-d data acquisition by rainbow range finder,'' in
  \emph{[1990] Proceedings. 10th International Conference on Pattern
  Recognition}, vol.~1.\hskip 1em plus 0.5em minus 0.4em\relax IEEE, 1990, pp.
  309--313.

\bibitem{Mirdehghan_2018_CVPR}
P.~Mirdehghan, W.~Chen, and K.~N. Kutulakos, ``Optimal structured light Ã la
  carte,'' in \emph{The IEEE Conference on Computer Vision and Pattern
  Recognition}, June 2018.

\bibitem{gupta2018geometric}
M.~Gupta and N.~Nakhate, ``A geometric perspective on structured light
  coding,'' in \emph{Proceedings of the European Conference on Computer Vision
  (ECCV)}, 2018, pp. 87--102.

\bibitem{zhang20173d}
Y.~Zhang, M.~Ye, D.~Manocha, and R.~Yang, ``3d reconstruction in the presence
  of glass and mirrors by acoustic and visual fusion,'' \emph{IEEE transactions
  on pattern analysis and machine intelligence}, vol.~40, no.~8, pp.
  1785--1798, 2017.

\bibitem{padilla2005advancements}
D.~Padilla and P.~Davidson, ``Advancements in sensing and perception using
  structured lighting techniques: An ldrd final report,'' 2005.

\bibitem{wang2016dual}
J.~Wang, A.~C. Sankaranarayanan, M.~Gupta, and S.~G. Narasimhan, ``Dual
  structured light 3d using a 1d sensor,'' in \emph{European Conference on
  Computer Vision}.\hskip 1em plus 0.5em minus 0.4em\relax Springer, 2016, pp.
  383--398.

\bibitem{bartels2019agile}
J.~R. Bartels, J.~Wang, W.~Whittaker, S.~G. Narasimhan \emph{et~al.}, ``Agile
  depth sensing using triangulation light curtains,'' in \emph{Proceedings of
  the IEEE International Conference on Computer Vision}, 2019, pp. 7900--7908.

\bibitem{GuptaA}
M.~Gupta, A.~Agrawal, and A.~Veeraraghavan, ``A practical approach to 3d
  scanning in the presence of interreflections, subsurface scattering and
  defocus,'' \emph{International Journal of Computer Vision}, vol. 102, no.
  1-3, pp. 33--55, 2013.

\bibitem{nayar2006fast}
S.~K. Nayar, G.~Krishnan, M.~D. Grossberg, and R.~Raskar, ``Fast separation of
  direct and global components of a scene using high frequency illumination,''
  in \emph{ACM Transactions on Graphics}, vol.~25, no.~3.\hskip 1em plus 0.5em
  minus 0.4em\relax ACM, 2006, pp. 935--944.

\bibitem{xu2007robust}
Y.~Xu and D.~G. Aliaga, ``Robust pixel classification for 3d modeling with
  structured light,'' in \emph{Proceedings of Graphics Interface 2007}, 2007,
  pp. 233--240.

\bibitem{gu2011multiplexed}
J.~Gu, T.~Kobayashi, M.~Gupta, and S.~K. Nayar, ``Multiplexed illumination for
  scene recovery in the presence of global illumination,'' in \emph{2011
  International Conference on Computer Vision}.\hskip 1em plus 0.5em minus
  0.4em\relax IEEE, 2011, pp. 691--698.

\bibitem{chen2008modulated}
T.~Chen, H.-P. Seidel, and H.~P. Lensch, ``Modulated phase-shifting for 3d
  scanning,'' in \emph{Proceedings of the IEEE Conference on Computer Vision
  and Pattern Recognition}.\hskip 1em plus 0.5em minus 0.4em\relax IEEE, 2008,
  pp. 1--8.

\bibitem{gupta2012micro}
M.~Gupta and S.~K. Nayar, ``Micro phase shifting,'' in \emph{Proceedings of the
  IEEE Conference on Computer Vision and Pattern Recognition}.\hskip 1em plus
  0.5em minus 0.4em\relax IEEE, 2012, pp. 813--820.

\bibitem{moreno2015embedded}
D.~Moreno, K.~Son, and G.~Taubin, ``Embedded phase shifting: Robust phase
  shifting with embedded signals,'' in \emph{Proceedings of the IEEE Conference
  on Computer Vision and Pattern Recognition}, 2015, pp. 2301--2309.

\bibitem{cronie2019coordination}
H.~S. Cronie, ``Coordination of multiple structured light-based 3d image
  detectors,'' Apr.~9 2019, uS Patent 10,257,498.

\bibitem{wu2020freecam3d}
Y.~Wu, V.~Boominathan, X.~Zhao, J.~T. Robinson, H.~Kawasaki,
  A.~Sankaranarayanan, and A.~Veeraraghavan, ``Freecam3d: Snapshot structured
  light 3d with freely-moving cameras,'' in \emph{European Conference on
  Computer Vision}.\hskip 1em plus 0.5em minus 0.4em\relax Springer, 2020, pp.
  309--325.

\bibitem{zhang2019causes}
Y.~Zhang, D.~L. Lau, and Y.~Yu, ``Causes and corrections for bimodal multi-path
  scanning with structured light,'' in \emph{Proceedings of the IEEE Conference
  on Computer Vision and Pattern Recognition}, 2019, pp. 4431--4439.

\bibitem{zhang2021sparse}
Y.~Zhang, D.~L. Lau, and D.~Wipf, ``Sparse multi-path corrections in fringe
  projection profilometry,'' in \emph{Proceedings of the IEEE Conference on
  Computer Vision and Pattern Recognition}, 2021.

\bibitem{comm_textbook}
J.~G. Proakis and M.~Salehi, \emph{Digital communications}.\hskip 1em plus
  0.5em minus 0.4em\relax McGraw-hill New York, 2001, vol.~4.

\bibitem{bch_code1}
A.~Hocquenghem, ``Codes correcteurs d'erreurs,'' \emph{Chiffers}, vol.~2, pp.
  147--156, 1959.

\bibitem{golay_code}
M.~J. Golay, ``Notes on digital coding,'' \emph{Proc. IEEE}, vol.~37, p. 657,
  1949.

\bibitem{turbo}
J.~Hagenauer, E.~Offer, and L.~Papke, ``Iterative decoding of binary block and
  convolutional codes,'' \emph{IEEE Transactions on information theory},
  vol.~42, no.~2, pp. 429--445, 1996.

\bibitem{ldpc}
R.~Gallager, ``Low-density parity-check codes,'' \emph{IRE Transactions on
  information theory}, vol.~8, no.~1, pp. 21--28, 1962.

\bibitem{lidar_cdma}
G.~Kim and Y.~Park, ``Lidar pulse coding for high resolution range imaging at
  improved refresh rate,'' \emph{Optics express}, vol.~24, no.~21, pp.
  23\,810--23\,828, 2016.

\bibitem{sagawa2017illuminant}
R.~Sagawa and Y.~Satoh, ``Illuminant-camera communication to observe moving
  objects under strong external light by spread spectrum modulation,'' in
  \emph{Proceedings of the IEEE Conference on Computer Vision and Pattern
  Recognition}, 2017, pp. 5097--5105.

\bibitem{li2021error}
J.~Li, J.~Guan, H.~Du, and J.~Xi, ``Error self-correction method for phase jump
  in multi-frequency phase-shifting structured light,'' \emph{Applied Optics},
  vol.~60, no.~4, pp. 949--958, 2021.

\bibitem{porras2017error}
R.~Porras-Aguilar, K.~Falaggis, and R.~Ramos-Garcia, ``Error correcting
  coding-theory for structured light illumination systems,'' \emph{Optics and
  Lasers in Engineering}, vol.~93, pp. 146--155, 2017.

\bibitem{ecc_textbook}
S.~Lin and D.~J. Costello, \emph{Error control coding}.\hskip 1em plus 0.5em
  minus 0.4em\relax Prentice hall New York, 2001, vol.~2, no.~4.

\bibitem{rm_code}
I.~S. Reed, ``A class of multiple-error-correcting codes and the decoding
  scheme,'' Massachusetts Inst of Tech Lexington Lincoln Lab, Tech. Rep., 1953.

\bibitem{geng2011structured}
J.~Geng, ``Structured-light 3d surface imaging: a tutorial,'' \emph{Advances in
  Optics and Photonics}, vol.~3, no.~2, pp. 128--160, 2011.

\bibitem{list_decode1}
P.~Elias, ``Error-correcting codes for list decoding,'' \emph{IEEE Transactions
  on Information Theory}, vol.~37, no.~1, pp. 5--12, 1991.

\bibitem{med_filter}
T.~Huang, G.~Yang, and G.~Tang, ``A fast two-dimensional median filtering
  algorithm,'' \emph{IEEE transactions on acoustics, speech, and signal
  processing}, vol.~27, no.~1, pp. 13--18, 1979.

\bibitem{MIFDB16}
N.~Mayer, E.~Ilg, P.~H{\"a}usser, P.~Fischer, D.~Cremers, A.~Dosovitskiy, and
  T.~Brox, ``A large dataset to train convolutional networks for disparity,
  optical flow, and scene flow estimation,'' in \emph{IEEE International
  Conference on Computer Vision and Pattern Recognition}, 2016.

\bibitem{sl_encodscopy}
C.~Schmalz, F.~Forster, A.~Schick, and E.~Angelopoulou, ``An endoscopic 3d
  scanner based on structured light,'' \emph{Medical image analysis}, vol.~16,
  no.~5, pp. 1063--1072, 2012.

\bibitem{xu2009adaptive}
Y.~Xu and D.~Aliaga, ``An adaptive correspondence algorithm for modeling scenes
  with strong interreflections.'' \emph{IEEE Transactions on Visualization and
  Computer Graphics}, vol.~15, no.~3, pp. 465--480, 2009.

\bibitem{matsuda2015mc3d}
N.~Matsuda, O.~Cossairt, and M.~Gupta, ``Mc3d: Motion contrast 3d scanning,''
  in \emph{IEEE International Conference on Computational Photography}.\hskip
  1em plus 0.5em minus 0.4em\relax IEEE, 2015, pp. 1--10.

\bibitem{wang2018programmable}
J.~Wang, J.~Bartels, W.~Whittaker, A.~C. Sankaranarayanan, and S.~G.
  Narasimhan, ``Programmable triangulation light curtains,'' in
  \emph{Proceedings of the European Conference on Computer Vision}, 2018, pp.
  19--34.

\bibitem{ancha2020active}
S.~Ancha, Y.~Raaj, P.~Hu, S.~G. Narasimhan, and D.~Held, ``Active perception
  using light curtains for autonomous driving,'' in \emph{European Conference
  on Computer Vision}.\hskip 1em plus 0.5em minus 0.4em\relax Springer, 2020,
  pp. 751--766.

\bibitem{n-ary_SL}
J.~L. Posdamer and M.~D. Altschuler, ``Surface measurement by space-encoded
  projected beam systems,'' \emph{Computer graphics and image processing},
  vol.~18, no.~1, pp. 1--17, 1982.

\bibitem{reed_solomon}
I.~S. Reed and G.~Solomon, ``Polynomial codes over certain finite fields,''
  \emph{Journal of the society for industrial and applied mathematics}, vol.~8,
  no.~2, pp. 300--304, 1960.

\end{thebibliography}

%\newpage
% \input{10_biography}

\vfill

\end{document}